\documentclass[10pt,twocolumn,letterpaper]{article}

\usepackage{wacv}
\usepackage{times}
\usepackage{epsfig}
\usepackage{graphicx}
\usepackage{amsmath}
\usepackage{amssymb}
\usepackage{booktabs}
\newcommand{\paragraphX}[1]{\vskip 0.2cm \noindent \textbf{#1} \hskip .1cm}
\usepackage{graphicx}
\usepackage{array}
\DeclareMathOperator*{\argmax}{arg\,max}

\usepackage{booktabs}
\usepackage{multirow}
\usepackage{dsfont}
\usepackage[mathscr]{eucal}
\usepackage{caption}
\usepackage{subcaption}
\usepackage{booktabs, multirow} 

\usepackage{wrapfig,lipsum}

\def\wacvPaperID{1365} 

\wacvalgorithmstrack   

\wacvfinalcopy 

\ifwacvfinal
\usepackage[breaklinks=true,bookmarks=false]{hyperref}
\else
\usepackage[pagebackref=true,breaklinks=true,colorlinks,bookmarks=false]{hyperref}
\fi

\begin{document}

\title{On Model Explanations with Transferable Neural Pathways}


\author{Xinmiao Lin\\
Rochester Institute of Technology\\
{\tt\small xl3439@rit.edu}
\and
Wentao Bao\\
Michigan State University\\
{\tt\small baowenta@msu.edu}
\and
Qi Yu\\
Rochester Institute of Technology\\
{\tt\small qi.yu@rit.edu}
\and
Yu Kong\\
Michigan State University\\
{\tt\small yukong@msu.edu}
}

\maketitle
\thispagestyle{empty}

\begin{abstract}
    Neural pathways as model explanations consist of a sparse set of neurons that provide the same level of prediction performance as the whole model. Existing methods primarily focus on accuracy and sparsity but the generated pathways may offer limited interpretability thus fall short in explaining the model behavior. In this paper, we suggest two interpretability criteria of neural pathways: (i) same-class neural pathways should primarily consist of class-relevant neurons; (ii) each instance's neural pathway sparsity should be optimally determined. To this end, we propose a \emph{Generative Class-relevant Neural Pathway} (GEN-CNP) model that learns to predict the neural pathways from the target model's feature maps. We propose to learn class-relevant information from features of deep and shallow layers such that same-class neural pathways exhibit high similarity. We further impose a faithfulness criterion for GEN-CNP to generate pathways with instance-specific sparsity. We propose to transfer the class-relevant neural pathways to explain samples of the same class and show experimentally and qualitatively their faithfulness and interpretability.  
\end{abstract}

\vspace{-1cm}

\section{Introduction}
\begin{figure}[t]
    \centerline{\includegraphics[width=0.45\textwidth]{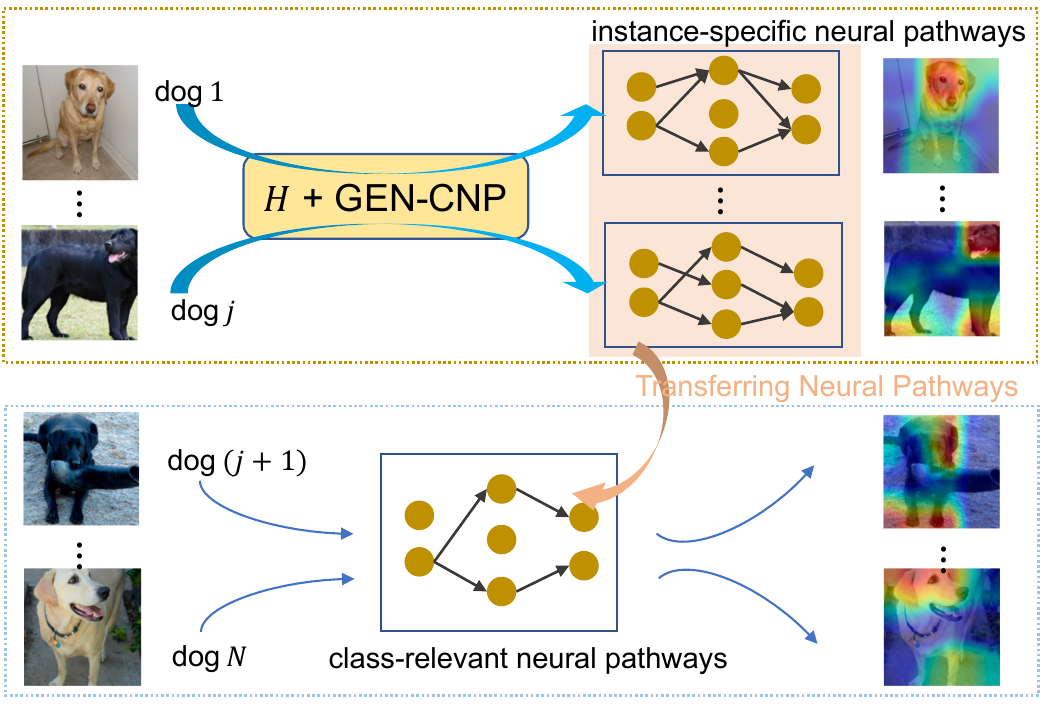}}
    \caption{\footnotesize{ \textbf{Transferable Neural Pathways.} Suppose that there are $N$ dog samples, $H$ is the target model and GEN-CNP is our model. \textbf{Top:} GEN-CNP generates instance-specific neural pathways explanations for $j$ dog instances. The class-relevant concepts revealed by the neural pathways are the head and legs and form the class-relevant neural pathways. \textbf{Bottom:} The class-relevant neural pathways are transferred to explain other dog samples (from $(j+1)$ to $N$) which again show the head and legs concepts. Class-relevant neural pathways provide a more global explanation for samples of the same class with consistent concepts revealed.}}
    \label{fig:concept}
\end{figure}

Model explanations through neural pathways should not only be sparse and faithful to the target model but also expected to reveal interpretable semantics \cite{PathwayGrad,dgr,cr-distillation}. Existing feature attribution methods \cite{cam,grad-cam,guided-backprop,integrated,extremalPerturbations,gradient,clrp} explain models through saliency maps, but neural pathways can additionally explain the internal states of these models. Therefore, neural pathway explanations can potentially satisfy a wider audience who has technical knowledge in DNNs and wishes to further understand the model's internal behavior in addition to the attention maps \cite{Ras2018ExplanationMI}. 

Existing methods produce neural pathways at instance-level but use a global sparsity for all samples, which is less faithful because some instances need more neurons and others less \cite{PathwayGrad,dgr,cr-distillation}. This gap inspires us to put forward the \textit{instance-specific interpretability} which requires the neural pathways to have optimal instance-level sparsity. Furthermore, we empirically discover that their neural pathways are highly variable for explaining same-class samples (see Figure \ref{fig:path_transfer_vis} in Section 4). This contradicts the observation that DNNs contain class-relevant neurons exclusively active for a given class \cite{glorot,cr-distillation,pmlr-v70-nagamine17a}. Interestingly, brain studies also show that groups of neurons are persistently fired for a category of faces/objects \cite{class_fir_1,class_fir_2,class_fir_3}. Inspired by these studies, we draw the \textit{class-wise interpretability} and suggest more class-relevant neurons to be included in the neural pathways. To summarize, we argue that the neural pathways explanations should not only be faithful at instance-level but also interpretable class-wise. When the explanations for a group of samples consistently reveal the class-relevant concepts, this could potentially benefit model designers to uncover bias in the model or aid model diagnostics.

We illustrate in Figure \ref{fig:concept} the benefit of class-wise interpretability: \textit{neural pathways transferability}. When instance-specific neural pathways select the class-relevant neurons which reveal the conceptual parts of the class, e.g., dog's head and legs, the aggregation of these neurons leads to the \textit{class-relevant neural pathways}. Then, by transferring the class-relevant neural pathways to explain other samples of the dog class, the explanations would also reveal the same class-relevant concepts. Neural pathways transferability allows more samples to be explained from a few and provides more consistent explanations across samples, i.e. explanations for explaining dogs always contain dog's head and legs. This helps people understand better the model's global internal states through ensembles of class-relevant neural pathways and such global explanations are much appreciated by the XAI community \cite{LIME,global_attributions,Ras2018ExplanationMI}. We later design a set of transferability experiments and empirically show that the class-relevant neural pathways can faithfully explain other samples of the same class, i.e. with accuracy comparable to the whole model. 

Hence, we propose a \textit{Generative Class-relevant Neural Pathways (GEN-CNP)} approach to learn to predict the neural pathways from the target model's feature maps. GEN-CNP first extracts the representative feature patterns from the feature maps, then predicts the importance of the feature patterns. In order to attribute higher importance to the class-relevant neurons, we leverage the observation that the features become more class-specific as the model goes deeper \cite{glorot,cr-distillation,pmlr-v70-nagamine17a} and propose a learning component that injects these class semantics of deeper layers into the earlier layers. Moreover, the training over the entire dataset allows GEN-CNP to distill better the class-relevant patterns. To convert the importance scores to the neural pathways, we introduce a decoding component with learnable sparsity constraint which allows the neural pathways to be of optimal sparsity at instance-level, hence satisfying the \textit{instance-specific interpretability}. We further impose a knowledge distillation loss to guide GEN-CNP towards generating faithful neural pathways satisfying class-wise and instance-specific interpretabilties. The components in GEN-CNP are built upon existing widely used modules such as convolution blocks which are easy to understand and do not introduce extra overhead for interpretability. We aim for this work to serve as a first step towards a more transferable interpretable explanation and hope our work can inspire future works to accomodate more complex target model architectures.

Our method differs from existing neural pathways explanation methods \cite{PathwayGrad,dgr,cr-distillation} in generating neural pathways. Existing methods generate importance scores, then prune the neurons based on a global sparsity which makes them resemble more to the model pruning \cite{prob_pruning,lecun_pruning} that are inadequate to improve model interpretability. We propose a learning-generation framework to fill out this critical gap, which offers both class-wise and instance-specific interpretabilities while conducting model pruning. We show through comprehensive experiments and visualizations that our neural pathways can achieve satisfactory faithfulness and interpretability. Our main contributions include:

\begin{itemize}
    \item a \textit{Generative Class-relevant Neural Pathways (GEN-CNP)} method to generate neural pathways explanations by exploring its internal inference states, which is more expressive than only using input features,
    \item the class-wise interpretability concept which provides an extra layer of interpretability by leveraging class-relevant neural pathways to explain other samples, and the concept of instance-specific interpretability which guarantees that each instance has its own unique sparsity in the generated neural pathways,
    \item a set of quantitative experiments to more accurately evaluate the faithfulness of the neural pathways along with qualitative analysis showing that the neural pathways are able to reveal meaningful semantics.
\end{itemize}

\section{Related Work}

\begin{figure*}[t!]
    \centerline{\includegraphics[width=1.\textwidth]{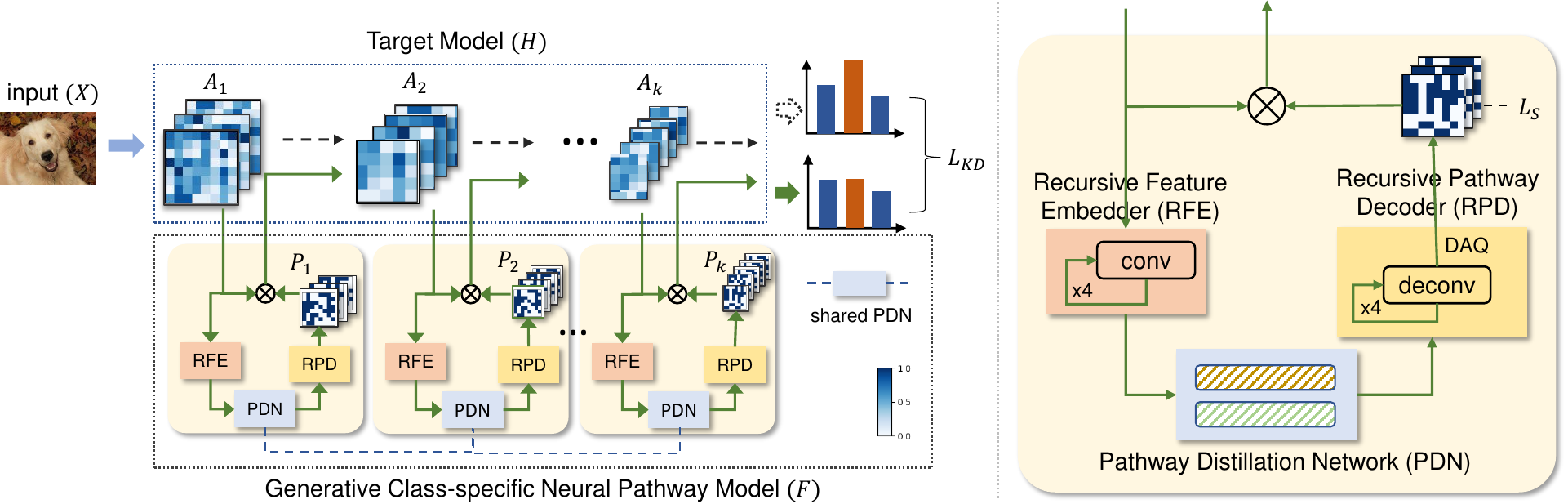}}
    \caption{\small 
    \footnotesize{ \textbf{GEN-CNP Overview.} The \textbf{left} illustrates the learning pipeline of GEN-CNP, the \textbf{right} shows the detailed layer-wise block of GEN-CNP. GEN-CNP takes as input the feature maps $A$ of the model $H$ and predicts the neural pathways $P$ given an instance $X$. The RFEs extract feature patterns that the PDN uses to predict importance scores. The RPDs then decode the importance scores to the binarized neural pathways. The knowledge distillation loss $L_{KD}$ between the prediction $H(X; P \otimes A)$ and the original prediction $H(X)$ with the sparsity loss $L_s$ for each $P_i$ guide GEN-CNP towards generating faithful and sparse neural pathways. }}
    \label{fig:gen-lp}
\end{figure*}

\paragraphX{Neural Pathways.} Neural pathways as explanations \cite{dgr,cr-distillation,PathwayGrad} find a sparse set of neurons in the original model. In addition to showing human-understandable explanations, neural pathways can, more importantly, explain the model's internal inference states. While existing feature attribution methods are mainly concerned with localizing the important input features for explaining a model's decision, they somehow neglect the model's interior. Existing neural pathways methods first compute the importance scores of neurons, then prune the neurons based on a sparsity for all samples. DGR \cite{dgr} computes the importance of neurons through iterative optimization for each instance. The authors in \cite{cr-distillation} propose to use (1) mean activation magnitude, (2) contribution index, and (3) AM visualization for filter/neuron importance scores computation. \cite{PathwayGrad} calculates the importance scores through Taylor approximation or IntGrad \cite{integrated}. As discussed previously, our method GEN-CNP proposes to predict the neural pathways for a target model with optimal instance-based firing sparsity and preserve the class-relevant knowledge in the target model.  

\paragraphX{Feature Attribution.} Feature attribution methods aim to localize the salient input features correlated with the model's decision. Class Activation Map (CAM) methods \cite{cam,grad-cam,grad-cam++,score-cam,ablation-cam,xgrad-cam,ucam,lfi-cam,relevance-cam} upsample a specific layer's feature map or a combination of feature maps for visualizing the model's salient features. The backpropagation-based methods \cite{lrp,clrp,deconvnet,excitation,deeplift,guided-backprop,gradient,integrated} calculate the saliency of input features by backpropagating the relevance scores from the output to the input. Each neuron's relevance score is often calculated using a combination of the gradients and the neuron values. Perturbation-based methods \cite{extremalPerturbations,smoothGrad,rise,FGVis,relex} calculate or optimize the importance of input features by perturbing them. Although the aforementioned methods produce straightforward explanations, they neglect the model's inference processes which can be addressed by neural pathways methods.

\paragraphX{Concept Attribution.} Concept attribution methods \cite{xprotonet,proto-tree,dan,nesy-xil,scg,this-that} produce explanations through visual concepts and often need to build an auxiliary model to approximate the target model. \cite{this-that} builds a prototype learning model and interprets the prediction as a combination of the prototypes. SCG \cite{scg} builds a structure concept graph where the nodes are the concepts and the edges are the contributions of concepts. \cite{proto-tree} constructs a decision tree with learnable visual concepts to approximate the target model. Similar to the feature attribution methods, concept attribution methods give interpretable explanations, but they still neglect the model's internal state and often need to learn new models or/with semantic concepts to approximate the target model. Thus, the explanations risk being more faithful to the new model than to the original model. The neural pathways instead give more faithful explanations because they can be thought of as a subnetwork of the original model, and no alteration or retraining of the original model is required.

\paragraphX{Model Pruning.} Model pruning, a technique in model compression, prunes the redundant weights of the model that are not sensitive to the performance \cite{pmlr-v139-wang21e,wang2021neural,pmlr-v139-wang21e}. Neural pathways are similar to model pruning because both give a sparse subnetwork. However, we differ from them in: (1) model pruning produces only one sparse subnetwork for all the samples, while our method generates one for each sample, (2) our goals are different, the sparse subnetwork of model pruning is still not interpretable, but our neural pathways have instance-specific and class-wise interpretability.

\section{Method}

\paragraphX{Overview.} We propose the Generative Class-relevant Neural Pathways (GEN-CNP) model to learn to predict the neural pathways from the target model's feature maps, shown in Figure \ref{fig:gen-lp}. In this paper, the target models are image recognition models. For each layer's feature maps, the Recursive Feature Embedders (RFEs) extract the representative feature patterns and embed them to the same resolution as other layers. The Pathway Distillation Network (PDN) predicts the importance scores of the feature patterns. The importance scores act as a precursor to the neural pathways, and the Recursive Pathway Decoders (RPDs) decode them into neural pathways. Neural pathways can be understood as binary masks to select the neurons, where a value of one indicates the neuron at the corresponding position is selected. In this section, we describe each component and its contribution towards achieving sparse and faithful neural pathways with class-wise and instance-specific interpretabilities.

\subsection{Generative Class-relevant Neural Pathways}
Denote the input as $X \in \mathbb{R}^{C \times H \times W}$, the target model to be explained as $H$, and our model GEN-CNP as $F$. Let the set of activation maps after the ReLU layers in $H$ be $A = \{A_1, A_2, ..., A_k \}$, $k$ is the number of ReLU layers in $H$. Note that $A_i$'s do not need to be of the same resolution, i.e. $A_i \in \mathbb{R}^{c_i \times h_i \times w_i}$ for $i = \{1, ..., k\}$. Denote the neural pathways as $P = \{P_1, ..., P_k\}$ where $P_i = [0, 1]^{c_i \times h_i \times w_i}$. GEN-CNP $F$ takes the feature maps $A$ and outputs the neural pathways $P$.

\begin{figure}[!t]
    \includegraphics[width=0.45\textwidth]{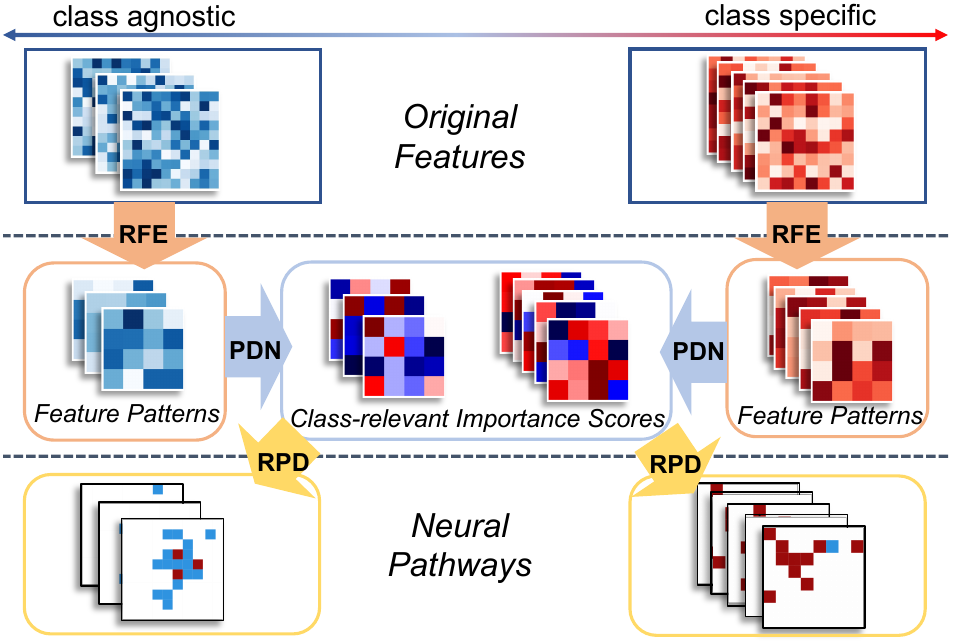}
    \centering
    \caption{\footnotesize\textbf{Conceptual Pipeline of GEN-CNP.} Taken two sets of feature maps as a conceptual example, one set from deeper layers (red), another from the earlier layers (blue). RFEs extract feature patterns from them and PDN learns to generate importance scores with class-relevant information. The RPDs decode the importance scores to the neural pathways.}
    \label{fig:pdn_concept}
\end{figure}



\paragraphX{RFE \& PDN.} The RFEs first extract the feature patterns from the feature maps $A$ in $H$ and the PDN predicts the importance scores of the feature patterns. To facilitate understanding, we illustrate the entire generation process from the feature maps to the neural pathways in Figure \ref{fig:pdn_concept}. 

For each activation maps $A_i$, the corresponding RFE is a recursive block composed of a convolution filter $g_i \in \mathbb{R}^{h_{g_i} \times w_{g_i}}$, a batch-normalization layer and an activation layer, as proposed in \cite{recursive-net}. Each RFE is recursively executed for $l_i$ iterations where $l_i = (h_i - h') / (h_{g_i}-1) = (w_i - w') / (w_{g_i}-1)$. Here, $h_{g_i}$ and $w_{g_i}$, convolution filter size, are chosen to be divisible by $(h_i-h')$ and $(w_i-w')$ and the $(h', w')$ are hyperparameters and shared for all RFEs. In our paper, we set $(h', w')$ to be the smallest feature map resolution. Thus, the outputs of all the layer-wise RFEs have the same resolution. For layers where the resolution of feature maps equals to $(h', w')$, we add padding to the convolution filter to make the number of iterations to be 1, more details in the supplement. A different batch normalization layer is used per iteration to mitigate the gradient explosion/vanishing problems \cite{recursive-net}. 

\textit{Motivation of Layer-wise RFE.} As motivated previously \cite{glorot,Goodfellow-et-al-2016,pmlr-v70-nagamine17a}, deeper layers encode class-relevant semantics, i.e. eyes/faces, while earlier layers encode low-level information, i.e. textures/edges. Using one RFE per layer allows the layer-wise class-specific/agnostic semantics to be separately preserved (second column in Figure \ref{fig:pdn_concept}). For more interpretable neural pathways, more representative feature patterns need to be preserved and convolutional filter is known to be more capable of capturing these patterns without losing locality information
compared to MLP, results in Table \ref{tab:ablation}. Layer-wise RFEs can also address the challenge of different resolutions across layers because they can be executed any amount of iterations to embed the features maps of all layers into the same resolution. Moreover, \cite{recursive-net} shows that a convolution block used recursively matches the embedding capability of a stack of convolution blocks which makes the RFEs computation- and resource-efficient.

\textit{PDN for Class-wise Interpretability.} Now, we propose the Pathway Distillation Network (PDN) to learn to map the feature patterns to the importance scores in all activation layers (second column to third column in Figure \ref{fig:pdn_concept}). In this paper, we use a MLP with activation of $z$ fully connected layers, note that MLP can be replaced with any backbone. 

We discuss now how the PDN contributes towards the class-wise interpretability. By design the PDN takes as input the features patterns of all layers, thus, it learns to extract the class-relevant semantics from deeper layers and injects them into the earlier layers. Also, these class-relevant semantics are more pronounced and easily distilled when the whole dataset is used for training. Note our goal is not to construct perfectly class-relevant neural pathways in all layers because this reduces the faithfulness of neural pathways to the original model. Instead, we aim for the neural pathways to reach a balance between interpretability and faithfulness, results shown in Figure \ref{fig:cIOU_vs_layer}.

In Figure \ref{fig:tsne_vis}, we represent the feature maps of the original model (top row) and the importance scores generated by PDN (bottom row) as embeddings and plot them using t-SNE \cite{tsne}. The first/second number measures the difference between the within-class/between-classes variance of PDN embeddings vs original feature embeddings. We first observe that the embeddings are grouped into more distinctive clusters/classes as the model goes deeper, which aligns with the observation in \cite{glorot,Goodfellow-et-al-2016,pmlr-v70-nagamine17a}. Then, the importance scores within-class are grouped much more tightly than the original features across layers which means that the class-relevant semantics successfully propagate from deeper layers to the earlier layers. Finally, the importance scores between-classes are also more dissimilar in layers 3 and 5, but still not overly class-relevant in layer 1, thus reaching a balance between faithfulness to the original model and the class-wise interpretability of neural pathways. Similarly, we show in Figure \ref{fig:cIOU_vs_layer} that our neural pathways contain more class-relevant neurons than the original features in the earlier layers, but are not overly biased towards them.  

\newcommand{\framewidth}{0.3\linewidth}
\newcolumntype{T}{>{\tiny}l}
\newcolumntype{H}{>{\Huge}l}

\begin{figure}[t]
\scriptsize
\centering
\renewcommand{\tabcolsep}{0pt} %
\renewcommand{\arraystretch}{0}
\begin{tabular}{>{\scriptsize}c c c c c c}
\parbox[l]{4mm}{\multirow{1}{*}[1.0em]{layer 1}} &
\parbox[l]{4mm}{\multirow{1}{*}[1.0em]{layer 3}} &
\parbox[l]{1cm}{\multirow{1}{*}[1.0em]{layer 5}} &
\parbox[l]{4mm}{\multirow{1}{*}[1.0em]{}}
\\
\parbox[c]{1mm}{\multirow{1}{*}[6.0em]{\rotatebox[origin=c]{90}{\tiny{original feature}}}} 
\includegraphics[width=\framewidth]{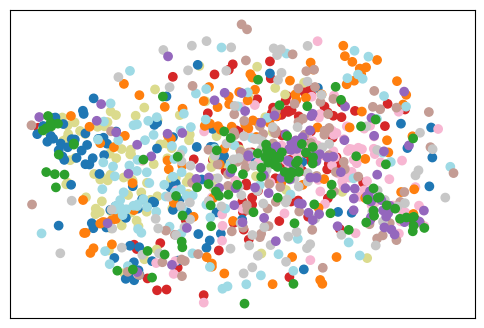} & 
\includegraphics[width=\framewidth]{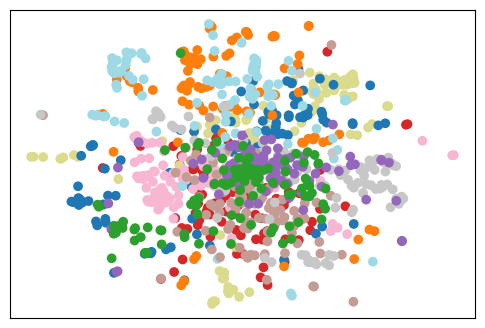} & 
\includegraphics[width=\framewidth]{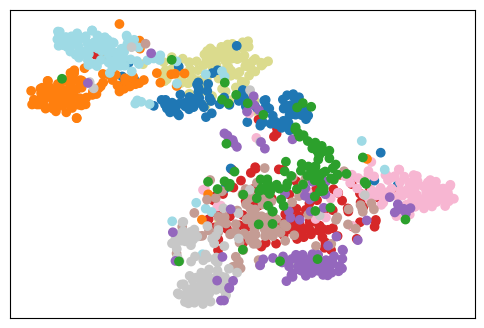} &  
\\
\parbox[c]{1mm}{\multirow{1}{*}[5.5em]{\rotatebox[origin=c]{90}{\tiny{PDN embeddings}}}} 
\includegraphics[width=\framewidth]{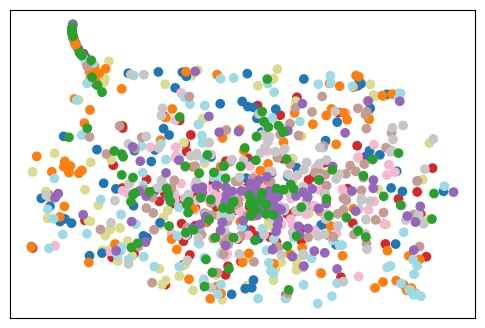} &
\includegraphics[width=\framewidth]{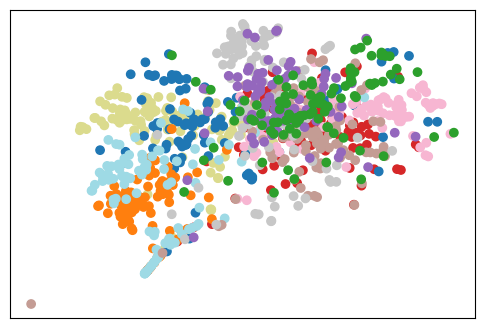} &
\includegraphics[width=\framewidth]{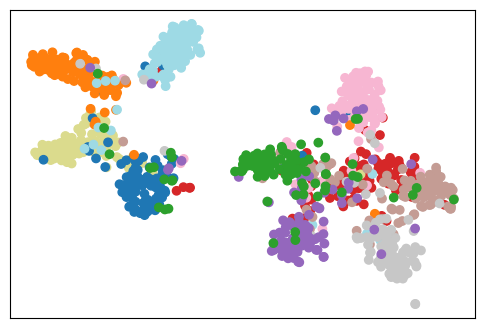} &
\\
\parbox[l]{2.1cm}{\multirow{1}{*}[0.0em]{\tiny{-87.52\%($\downarrow$);-25.66\%($\uparrow$)}}} &
\parbox[l]{2.4cm}{\multirow{1}{*}[0.0em]{\tiny{-10.33\%($\downarrow$);+17.08\%($\uparrow$)}}} &
\parbox[l]{2.2cm}{\multirow{1}{*}[0.0em]{\tiny{-13.23\%($\downarrow$);+0.34\%($\uparrow$)}}} &
\parbox[l]{1mm}{\multirow{1}{*}[0.0em]{}}
\\
\end{tabular}
\captionsetup{font=footnotesize,aboveskip=10pt}
\caption{t-SNE visualizations of the original features \textbf{(top)} vs the importance scores (PDN outputs) \textbf{(bottom)}. The bottom numbers are differences between ours and original of: the embeddings variance within-class (first number), and the embeddings variance between classes (second number). The results are on AlexNet \cite{alexnet} and CIFAR-10 test set \cite{cifar10} with colors representing the classes. More results in the supplement.}
\label{fig:tsne_vis}
\end{figure}
\paragraphX{RPD.} The importance scores are not readily the neural pathways because they are on the low-dimensional space and not binarized (third column in Figure \ref{fig:pdn_concept}). To this purpose we propose the layer-wise Recursive Pathway Decoders (RPD), consisting of the recursive blocks as in layer-wise RFEs and Distance Aware Quantization (DAQ) blocks. 

Similar to the layer-wise RFEs, a recursive block decodes the $i$-th layer's importance scores to the same resolution as $A_i$. The recursive blocks are configured similarly to the ones in RFEs except that the convolution block is replaced with the transposed convolution block. Let the outputs of the recursive blocks be $D = \{D_1, ..., D_k\}$, where $D_i \in \mathbb{R}^{c_i \times h_i \times w_i}$. We then propose to integrate a Distance Aware Quantization (DAQ) in each layer to binarize $D_i$ to neural pathways\cite{daq}. DAQ is differentiable and aids GEN-CNP to generate sparse and faithful neural pathways. 
Specifically, we propose to quantize the real-valued $D_i$ by assigning the minmax normalized values of $D_i$ to their nearest binary integers (0 or 1):
\begin{equation}
\left\{
    \begin{split}
        \hat{D}_i &= (2^b-1) \frac{ \max(\min(D_i, l), u)  - l}{u-l}, \\
        P_i &= \frac{\phi(\hat{D}_i)}{2^b-1},
    \end{split}
\right.
\end{equation}
where $\phi$ is the soft assignment function which assigns the normalized values $\hat{D}_i$ to their nearest quantized values based on dynamic distance computation. $b$ is the number of quantized values, because the neural pathways are binary, $b=1$ by default in this paper, The $u$ and $l$ are the upper- and lower-bound hyperparameters same as in \cite{daq}.


\subsection{GEN-CNP Learning}
Because the ground truth of neural pathways explanations does not exist in practice, straightforward supervised learning of GEN-CNP is infeasible. We thus resort to the knowledge distillation learning with sparsity constraints for training the GEN-CNP \cite{kd_1,kd_2}. Knowledge distillation aims to learn a smaller model (neural pathways) from the output of a large model (the target model) such that faithfulness could be guaranteed without relying on the neural pathway ground truth. Specifically, we propose a knowledge distillation loss that penalizes the discrepancy between the target model's prediction and the prediction of neural pathways. We use the cross-entropy loss in this paper, as follows:
\begin{equation}
    \begin{split}
       L_{\text{KD}} &= L_{\text{CE}}\left({H(X; P \otimes A), H(X)}\right) .
        \label{eq:lr}
    \end{split}
\end{equation}
$H(X)$ is the original prediction, and $H(X; P \otimes A)$ is the prediction given by the subnetwork $(P \otimes A)$. 

To achieve sparse neural pathways, $l_0$-norm would be the most straightforward sparsity constraint, i.e., $L_{S} = \Vert P \Vert_0$. However, optimizing the $l_0$-norm is shown to be NP-hard \cite{Natarajan1995SparseAS}. Thus, we use the $l_2$-norm and the total loss is:
\begin{equation}
    \begin{split}
        L = \alpha L_{KD} + \beta \Vert P \Vert^2_2 ,
        \label{eq:ltot}
    \end{split}
\end{equation}
where $\alpha$ and $\beta$ are the tunable hyperparameters and the details are in the supplement.

\textit{Towards Instance-specific Interpretability.} Recall our motivation for instance-specific interpretability to bridge the gap between instance-level neural pathways and global sparsity in the existing neural pathways methods \cite{PathwayGrad,cr-distillation,dgr}. The sparsity constraint guides the RPDs to binarize importance scores to sparse neural pathways, and the knowledge distillation loss constrains the neural pathways to be faithful to the original model. In Table \ref{tab:ablation} third row where we use a global sparsity for all samples, the faithfulness to the original model has decreased. The knowledge distillation loss also constrains the PDN to not only attribute importance scores to class-relevant neurons which if selected, would make the neural pathways less faithful. 

\section{Experiments}

\paragraphX{Datasets \& Models.} We perform experiments on two models: AlexNet \cite{alexnet} and VGG-11 with batch normalization \cite{vgg}. The datasets used are CIFAR-10 \cite{cifar10} which is an image dataset of 10 classes and 60,000 images and ImageNet \cite{imagenet} which is a large-scale image dataset of 1,000 classes. We test on the entire test set of CIFAR-10 and validation set of ImageNet. Due to limited computational resources, we downsize the ImageNet dataset to the resolution as the CIFAR-10. The hyperparameters used for each model and dataset and the training details are in the supplement.

\paragraphX{Evaluation Metrics.} Below are the evaluation metrics used in this paper.
\begin{itemize}
    \item \textbf{Accuracy.} Accuracy measures the average number of samples correctly classified when the feature maps are multiplied with the neural pathways .
    
    
    \item \textbf{Mean Increase in Confidence (mIC).} We introduce a metric similar to ICr in \cite{grad-cam++} to measure the mean increase in confidence for correctly classified samples with increased confidence, $\text{mIC} = \sum_i^n (\hat{y}_i - y_i) \times \mathds{1}_{ \{(\hat{y}_i > y_i) \wedge \hat{c}_i = c_i\}} / n$. ICr results are in the supplement.
    
    \item \textbf{Mean Decrease in Confidence (mDC).} Similarly to mIC, $\text{mDC} = \sum_i^n (y_i - \hat{y}_i ) \times \mathds{1}_{ \{(\hat{y}_i < y_i) \wedge \hat{c}_i = c_i\}} / n$.
    
    \item \textbf{Average Class IOU (acIOU).} We introduce the acIOU metric to quantitatively measure the similarity between neural pathways for samples in one class, averaged across all classes, $\text{acIOU} = \sum_{c \ \in C} \sum_{x_i, x_j, x_i \neq x_j}^{|X_c|} \frac{|P_{x_i} \cap P_{x_j}|}{2n|P_{x_i} \cup P_{x_j}|}$.
    
\end{itemize}

\paragraphX{Comparison Methods.} Existing neural pathways explanation methods use a sparsity for all samples, during comparison, the sparsity used for them would be the average sparsity of our neural pathways. The baseline methods are NeuronMCT and Neuron Grad \cite{PathwayGrad} which use Taylor approximation and IntGrad \cite{integrated} to compute the importance scores. DGR and DGR\_R \cite{dgr} instead iteratively optimize the importance scores of neurons and they differ in their initialization, DGR initializes with ones while DGR\_R initializes with random values. We also compare with the magnitude pruning which uses the neuron values as importance scores. \cite{PathwayGrad} also proposes the \textit{Greedy Pruning} which greedily selects the least important neurons leading to the correct prediction. Another benchmark is the \textit{Random} method which randomly selects neurons to keep. 

\subsection{Quantitative Analysis}

Table \ref{tab:imagenet-fait} summarizes the faithfulness performance of the neural pathways on ImageNet \cite{imagenet}, more results are in the supplement. Generally, our neural pathways are shown to induce higher model confidence than other methods. The first reason is that the neurons which influence adversely the model's decision are not selected by our neural pathways. Another reason is the instance-specific interpretability achieved by our method. In other words, our neural pathways can select more or fewer neurons to be more faithful to the original model while existing methods have one firing sparsity for all samples \footnote{Firing sparsity is the same as the sparsity in previous sections.}. We show in the supplement that when we use a global firing sparsity as the baseline methods, the neural pathways are less faithful to the model. 

\begin{table}[!tp]\centering
\scriptsize
\setlength{\tabcolsep}{0.6mm}
\caption{\footnotesize{ Results (\%) on AlexNet \cite{alexnet} and VGG-11 \cite{vgg} models for ImageNet \cite{imagenet}. The test accuracies of AlexNet and VGG-11 are $99.93\%$ and $99.92\%$. The Accuracy metric is for reference only. More results are in the supplement. }}
\begin{tabular}{l|ccc|ccc}
\hline
\multirow{2}{*}{Method} & \multicolumn{3}{c|}{AlexNet} & \multicolumn{3}{c}{VGG11} \\
\cline{2-7}
& mIC ($\uparrow$) & mDC ($\downarrow$) & Acc. ($\uparrow$) & mIC ($\uparrow$) & mDC ($\downarrow$) & Acc. ($\uparrow$) \\
\hline
Greedy \cite{PathwayGrad} &1.00 &46.10 &8.39 &0.00 &91.13 &3.76 \\
Random &0.00 &96.79 &0.17 &0.00 &95.32 &0.12 \\
Original Activation \cite{lecun_pruning} &0.00 &25.17 &26.07 &0.00 &92.27 &96.49 \\
\hline
NeuronMCT \cite{PathwayGrad} &0.00 &92.83 &89.47 &0.00 &95.31 &99.90 \\
NeuronGrad \cite{PathwayGrad} &0.00 &92.60 &89.81 &0.00 &95.29 &99.90 \\
DGR \cite{dgr} &1.07 &51.99 &97.93 &0.00 &94.20 & 99.95 \\
DGR\_R \cite{dgr} &1.15 &79.01 &97.89 &0.00 &94.66 &99.94 \\
\textbf{GEN-CNP (ours)} &\textbf{3.09} &\textbf{9.71} &99.18 &\textbf{1.06} &\textbf{23.80} &99.74 \\
\hline
\end{tabular}
\label{tab:imagenet-fait}
\end{table}


\begin{figure*}[!tp]
\captionsetup[subfigure]{font=scriptsize,labelfont=scriptsize}
\begin{subfigure}{.33\textwidth}
  \centering
  \includegraphics[width=1\linewidth]{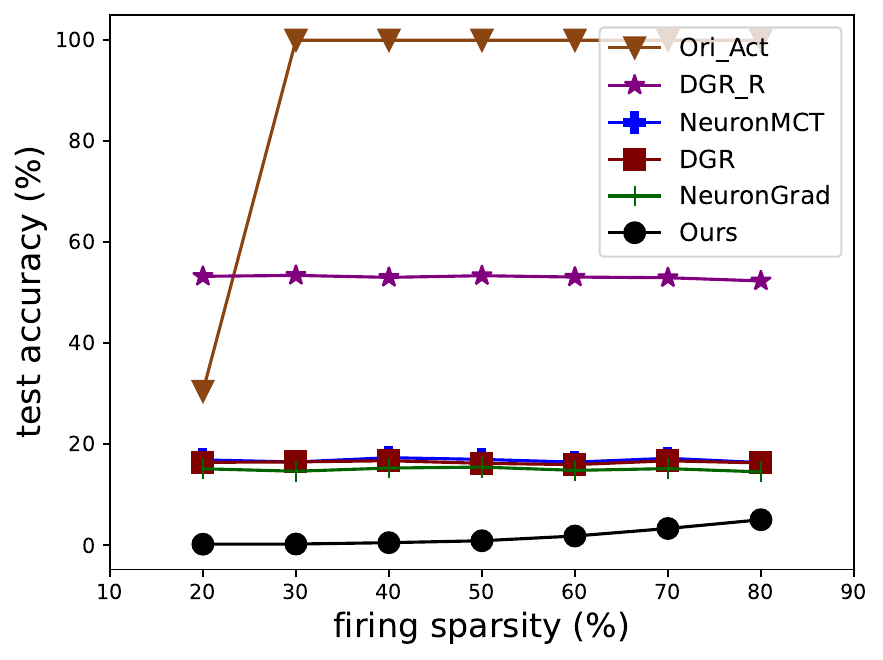}
  \subcaption{ROAP on AlexNet ImageNet}
  \label{fig:roap}
\end{subfigure}
\begin{subfigure}{.33\textwidth}
  \centering
  \includegraphics[width=1\linewidth]{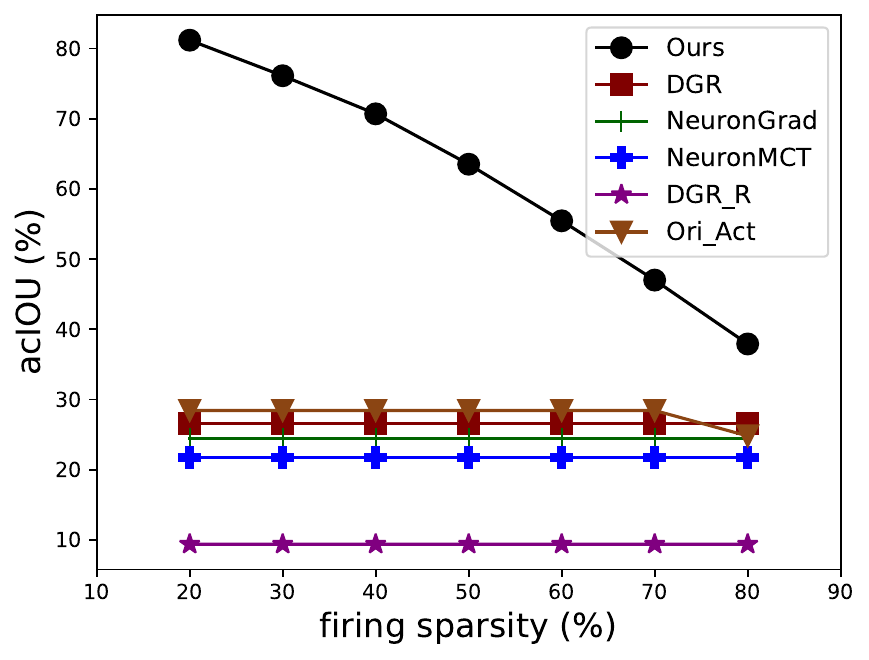}
  \subcaption{acIOU vs firing sparsity}
  \label{fig:cIOU_vs_firs}
\end{subfigure}%
\begin{subfigure}{.33\textwidth}
  \centering
  \includegraphics[width=1\linewidth]{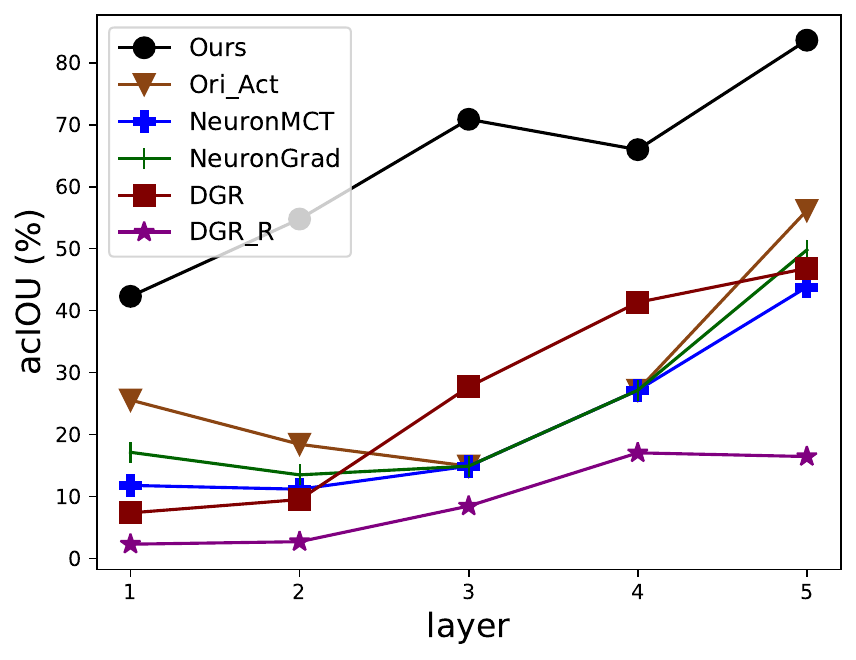}
  \subcaption{acIOU for each layer}
  \label{fig:cIOU_vs_layer}
\end{subfigure}
\caption{\footnotesize{ \textbf{(a).} ROAP experiment on AlexNet \cite{alexnet} and ImageNet \cite{imagenet}. \textbf{Lower} increase is \textbf{better}. Higher firing sparsity means that neural pathways contain fewer neurons. \textbf{(b).} acIOU vs firing sparsity for AlexNet \cite{alexnet} on CIFAR-10 \cite{cifar10}. \textbf{(c).} acIOU for each layer of AlexNet on CIFAR-10. More results are in the supplement. }}
\label{fig:transfer}
\end{figure*}


\paragraphX{RemOve-And-Predict (ROAP).} RemOve And Retrain (ROAR) \cite{roar} removes the salient input features and retrains the model on the remaining features. Deletion metric \cite{rise} deletes the localized input features and measures the accuracy on the remaining features. Similarly, we propose the RemOve and Predict (ROAP) benchmark to assess the faithfulness of the neural pathways. We assume that more faithful neural pathways, if removed, will cause the model's performance to drop more significantly. 

Figure \ref{fig:roap} shows that compared to baselines, when the model contains more neurons, it still has low accuracy using the remaining neurons. This shows that our neural pathways are more faithful because they contain the most important neurons for the model's decisions. Note that the baselines NeuronMCT, NeuronGrad, DGR have similar test accuracy patterns and could suggest that the computed neurons have similar important scores although the methods of computing them are different. Also, the pruning method \textit{Original Activation} has reached almost 100\% test accuracy which indicates that the neuron magnitude is not equivalent to the importance scores as we often believe \cite{guided-backprop,lrp}.



\begin{table}
\scriptsize
\setlength{\tabcolsep}{0.5cm}
\caption{\footnotesize{ \textbf{GEN-CNP Components Ablation Experiments}. 
Results (\%) are on AlexNet \cite{alexnet} and CIFAR-10 \cite{cifar10}}}
\begin{tabular}{ >{\centering\arraybackslash}m{1.5cm}| c  c  c  c }
\hline
Method & mIC ($\uparrow$) & mDC ($\downarrow$) & Acc. ($\uparrow$) \\ \hline
w/o RFEs & 0.87  & 51.75 & 73.43  \\ 
l.-w. PDNs  & 0.00 & 43.54 & 76.69 \\ 
w/o DAQ & 6.31 & 17.70 & 76.97 \\ \hline
\textbf{Ours} &\textbf{14.67} &\textbf{14.40} &\textbf{77.12} \\
\hline
\end{tabular}
\label{tab:ablation}
\end{table}

\paragraphX{Ablation Studies.} As motivated in the methodology section, we experimentally show how each component in GEN-CNP, namely RFE, PDN, RPD, acts towards more faithful neural pathways in Table \ref{tab:ablation}. The first model (w/o REFs) consists of layer-wise PDNs without RFEs and directly predicts the importance scores for the feature maps. The second model (l.-w. PDNs) is to replace the shared PDN with layer-wise PDNs. The third model (w/o DAQ) is GEN-CNP without DAQ which generates importance scores instead of neural pathways with dynamic instance-specific sparsity. 

Using one PDN per layer (second row) has lower accuracy than our method shows that using convolutional filters in RFEs is crucial to preserve the representative feature patterns which improve faithfulness as well. The models (first and second row) do not allow class-relevant semantics to be shared among layers and show lower faithfulness than our method in mIC and mDC. This implies that class-relevant neural pathways are faithful explanations while being more interpretable. The decrease in accuracy for the third models is because a global sparsity is used for all samples and shows that instance-specific interpretability does lead to more faithful neural pathways as we have speculated and motivated previously.


\subsection{Transferability Experiments}
In the introduction, we motivated the importance of class-relevant neural pathways and described in the method section how GEN-CNP achieves it. We now experimentally demonstrate that they not only reveal class-relevant semantics but also can be transferred to explain faithfully other samples of the same class.

\begin{figure*}[!tp]
\captionsetup[subfigure]{font=scriptsize,labelfont=scriptsize}
\begin{subfigure}{.33\textwidth}
  \centering
  \includegraphics[width=1.1\linewidth]{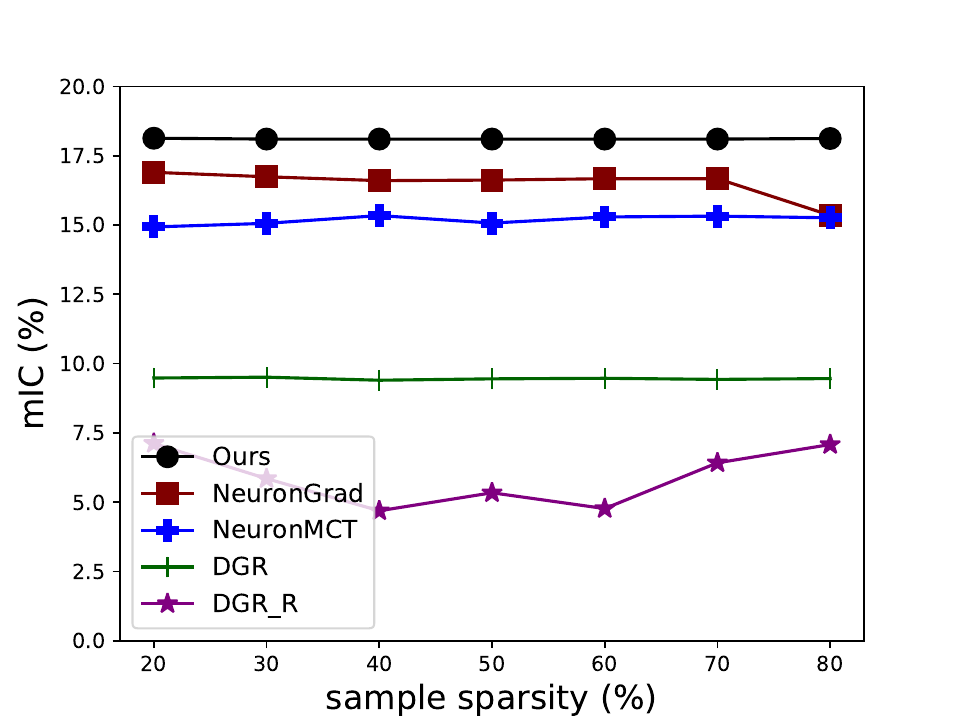}
  \caption{mIC vs sample sparsity $\epsilon_{ss}$}
  \label{fig:aic_vs_ss}
\end{subfigure}%
\begin{subfigure}{.33\textwidth}
  \centering
  \includegraphics[width=1.1\linewidth]{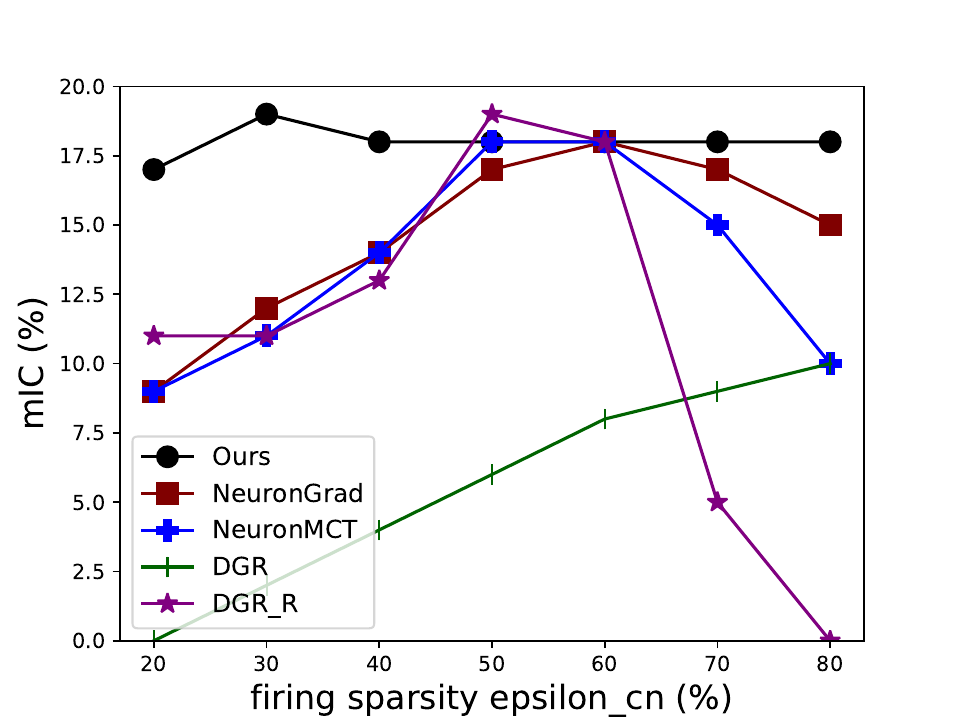}
  \caption{mIC vs $\epsilon_{cn}$}
  \label{fig:aic_vs_ms}
\end{subfigure}
\begin{subfigure}{.33\textwidth}
  \centering
  \includegraphics[width=1.1\linewidth]{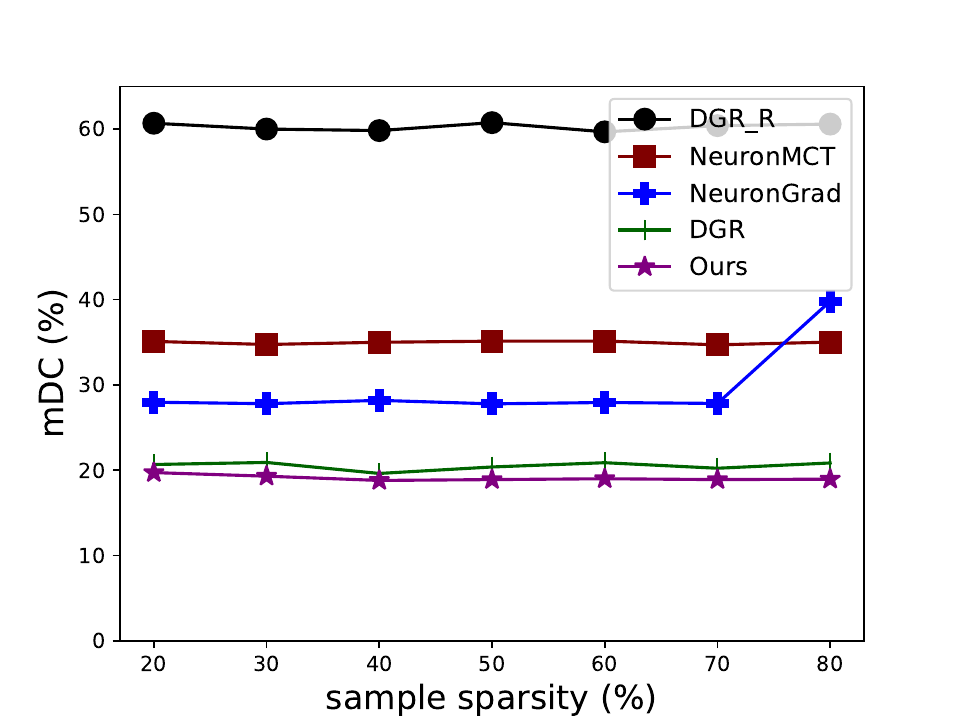}
  \caption{mDC vs $\epsilon_{ss}$}
  \label{fig:adc_vs_ss}
\end{subfigure}
\caption{\footnotesize{ \textbf{Transferability Experiments on AlexNet \cite{alexnet} and CIFAR-10 \cite{cifar10}} \textbf{(a).} mIC vs sample sparsity $\epsilon_{ss}$ given a firing sparsity of 95\% and class-relevant sparsity of 70\%. \textbf{(b).} mIC vs class-relevant sparsity given the firing sparsity of 95\% and sample sparsity of 60\%. \textbf{(c).} mDC vs sample sparsity $\epsilon_{ss}$ given a fixed firing sparsity of 95\% and class-relevant sparsity of 70\%. More results are in the supplement. }}
\label{fig:transfer}
\vspace{-0.5cm}
\end{figure*}

\vspace{-0.2cm}
\paragraphX{Average Class IOU.} We report the results on acIOU in Table \ref{tab:ciou}. We observe that GEN-CNP achieves higher acIOU compared to the baselines which means higher overlap between neural pathways of a given class. Given also they are faithful (Table \ref{tab:imagenet-fait}) which implies that the common neurons among the neural pathways are class-relevant. In other words, our neural pathways contain more class-relevant neurons. Also, the subnetwork that aggregates all the class-relevant neurons is more compact and could potentially inspire future research in developing lightweight neural networks that are interpretable as well.

\begin{table}
\scriptsize
\caption{\footnotesize{ \textbf{Average Class IOU (acIOU) (\%)}. \textbf{Higher} is \textbf{better} means that the neural pathways of a given class contain more class-relevant neurons. More results are in the supplement. }}
\setlength{\tabcolsep}{0.3cm}
\begin{tabular}{l| c  c | c  c }
\hline
\multirow{2}{*}{Method} &\multicolumn{2}{c|}{CIFAR10} &\multicolumn{2}{c}{ImageNet} \\
\cline{2-5}
 &AlexNet &VGG-11 &AlexNet &VGG-11 \\ \hline
Ori. Act. \cite{lecun_pruning} & 1.78 &0.00  & 0.00  &8.09  \\
Greedy \cite{PathwayGrad} & 0.00 &0.00 &0.00 &0.00 \\
Random &0.00 &0.00  &0.00 &0.00 \\
\hline
NeuronMCT \cite{PathwayGrad} &9.85 &19.01 &18.71 &14.97 \\
NeuronGrad \cite{PathwayGrad} &10.25 &19.62 &18.68 &17.53 \\
DGR \cite{dgr} &15.64 &21.72 &19.36 &27.38 \\
DGR\_R \cite{dgr} & 8.95 & 19.09 & 19.40 & 7.33 \\
\hline
\textbf{GEN-CNP (ours)} & \textbf{17.99} & \textbf{40.24} & \textbf{59.96}  & \textbf{58.04} \\
\hline
\end{tabular}
\label{tab:ciou}
\end{table}


\paragraphX{Neural Pathways Analysis.} Figure \ref{fig:cIOU_vs_firs} shows our neural pathways can retain more class-relevant neurons compared to baselines even becoming sparser, which is not the case in the baselines and suggests that their neural pathways are highly variable, similar finding in Table \ref{tab:ciou}. Figure \ref{fig:cIOU_vs_layer} shows similar observation as Table \ref{tab:ciou} that our neural pathways are more class-relevant in general. As motivated in the methodology section, we aim for neural pathways to reach a balance between class-wise interpretability and faithfulness. Thus, ideal neural pathways should show an increasing trend of containing more class-relevant neurons as the model goes deeper, with class-agnostic/specific characteristics being preserved to some extent, as confirmed in both Figure \ref{fig:cIOU_vs_layer} and \ref{fig:tsne_vis}. 


\begin{figure*}[!t]
    \centerline{\includegraphics[width=.9\textwidth]{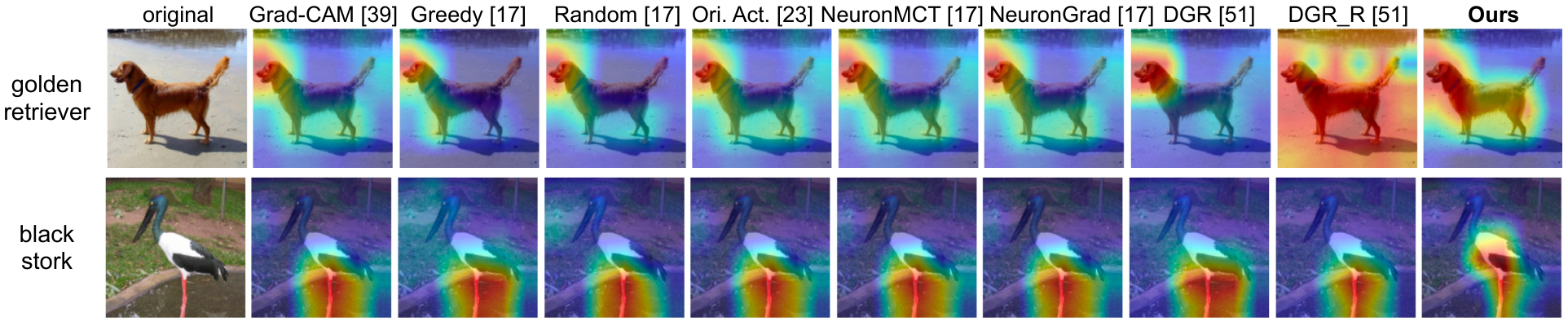}}
    \caption{\footnotesize{CAM Visualizations on ImageNet \cite{imagenet} using Grad-CAM. Our method compared to other methods seem to show less noise while focus more on the target animal, more results in the supplement.}}
    \label{fig:cam_image_vis}
\end{figure*}

\begin{figure*}[t]
    \centerline{\includegraphics[width=.9\textwidth]{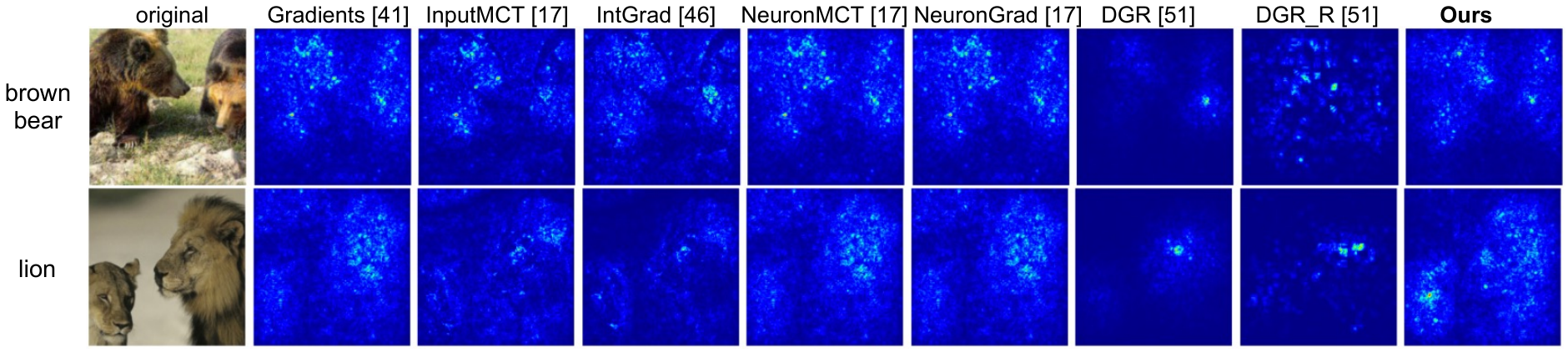}}
    \caption{\footnotesize{Neural Pathway Gradients Visualizations on ImageNet \cite{imagenet}. Our method compared to other methods seem to show less noise while have more salient features focus more on the target animal, more results in the supplement.}}
    \label{fig:pathgrad_image_vis}
\end{figure*}

\paragraphX{Transferability Settings.} As motivated in Figure \ref{fig:concept}, given the class-relevant neurons, we can extract the class-relevant neural pathways and transfer them to explain other samples of the same class. Let $X_c$ be the set of samples in class $c$, let $\hat{X}_c$ be the subset of $X_c$ given a \textit{sample sparsity} $\epsilon_{ss}$. For example, for $\epsilon_{ss} = 0.8$, $\hat{X}_c$ would contain 20\% of randomly chosen samples from $X_c$. Let $B = \frac{1}{|\hat{X}_c|} \sum_{x \in \hat{X}_c} \sum_{i}^k P_{i, (o, p, q)}(x)$ be the average firing sparsity rates for all neurons in the neural pathways $P$. Thus, given a class-relevant sparsity threshold $\epsilon_{cn}$, the class-relevant neural pathway is calculated as $P_c = \sum_{i}^k \mathds{1}_{ B_{\{i, (o, p, q)\} > \epsilon_{cn}}}$.


\paragraphX{Discussion.} Figure \ref{fig:transfer} gives the transferability experiments results for AlexNet \cite{alexnet} on CIFAR-10 \cite{cifar10}. Figure \ref{fig:aic_vs_ss}, \ref{fig:aic_vs_ms} and \ref{fig:adc_vs_ss} show that our class-relevant neural pathways are able to achieve higher faithfulness than the baseline methods consistently. We also observe that the mIC and mDC do not decrease when less samples are being used to compute the class-relevant neural pathways, i.e. lower $\epsilon_{ss}$, shown in Figure \ref{fig:aic_vs_ms} and Figure \ref{fig:adc_vs_ss}. This seems rather counter-intuitive because we expect the model to distill less class-relevant knowledge when fewer samples are used. However, the results suggest that the class-relevant neurons are equally relevant for 1 or many samples and our neural pathways can select these most class-relevant neurons regardless. Interestingly, Figure \ref{fig:aic_vs_ms} shows that the mIC of the baselines methods NeuronMCT \cite{PathwayGrad}, NeuronGrad \cite{PathwayGrad}, and DGR\_R \cite{dgr} first increases, then decreases when $\epsilon_{cn}$ continuously increases, this could suggest that when allowed to select more neurons, class-relevant neurons are not being assigned more importance scores than other neurons, hence not selected.

\begin{figure*}
    \centerline{\includegraphics[width=1.0\textwidth]{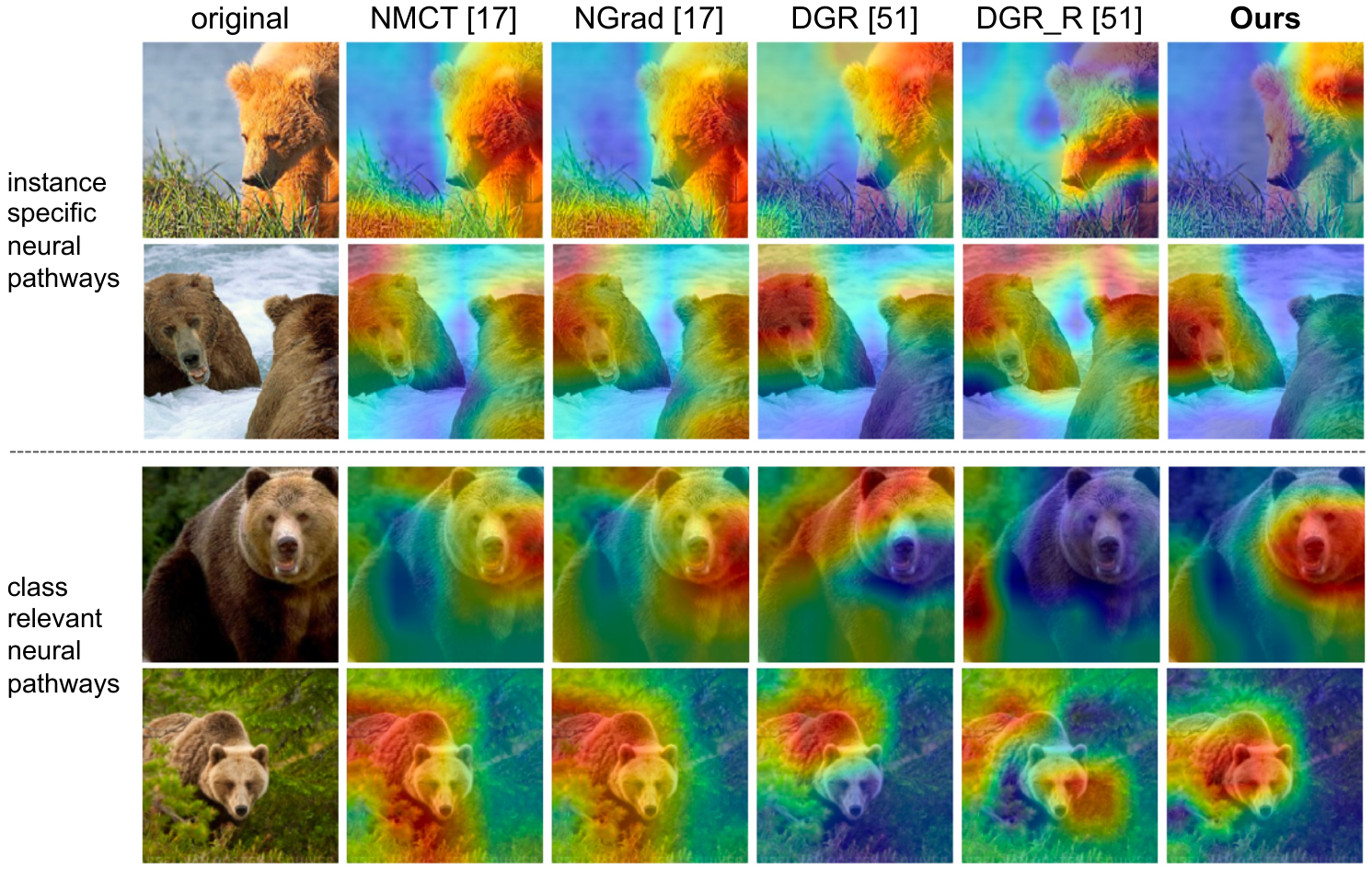}}
    \caption{\footnotesize{CAM Visualizations on ImageNet \cite{imagenet} using Grad-CAM. The class-relevant neural pathways when transferred, focus more tightly on the head and body of the brown bear, and less on the background.}}
    \label{fig:path_transfer_vis}
\end{figure*}


\paragraphX{Feature Attribution Visualizations.} In Figure \ref{fig:cam_image_vis}, we propose to visualize the concepts revealed in the deeper layers by the neural pathways and for this purpose, we integrate Grad-CAM \cite{grad-cam} on the neural pathways for AlexNet \cite{alexnet} on ImageNet \cite{imagenet}. Note that any CAM method can be used. The firing sparsity of neural pathways is 0.3. The visualizations of our neural pathways seem to focus more on the foreground animal/object and less on the background as compared to the baseline methods. We think that this could be due to our method being able to select more class-relevant neurons which are reflected to be class-relevant concepts, e.g. the body of black stork (second row of Figure \ref{fig:cam_image_vis}). In Figure \ref{fig:pathgrad_image_vis}, we visualize the gradients of the output to the input when only the gradients on the neural pathways are considered \cite{PathwayGrad}. Brighter pixels mean that the gradient values are higher which suggest their higher importance for a model's decision. Similarly, we observe that the saliency maps tend to be less noisy on our neural pathways than the baselines. 

In Figure \ref{fig:path_transfer_vis}, our instance-specific neural pathways (top part) reveal more clearly the concepts relevant to the brown bear such as the bear's head and body with less focus on the background. Thus, when transferring the class-relevant neural pathways to explain other samples of the brown bear, the same concepts are also revealed. For better visualization results and due to limited computation resources, we train GEN-CNP on a subset of 200 classes of ImageNet, more quantitative and qualitative results are in the supplement.


\section{Conclusion}
In this paper, we propose the Generative Class-relevant Neural Pathways (GEN-CNP) method for explaining image recognition models through neural pathways. We motivate the importance of class-wise and instance-specific interpretabilities of neural pathways in addition to their faithfulness and sparsity. We also propose the neural pathways transferability concept which transfers the class-relevant neural pathways to explain other samples of the same class. We believe that class-wise interpretability can constitute a step forward towards a more global explanation framework through neural pathways. 

{\small
\bibliographystyle{ieee_fullname}
\bibliography{egbib}

\begin{thebibliography}{10}\itemsep=-1pt

\bibitem{dan}
Moritz Böhle, Mario Fritz, and Bernt Schiele.
\newblock Convolutional dynamic alignment networks for interpretable
  classifications.
\newblock In {\em CVPR}, 2021.

\bibitem{grad-cam++}
Aditya Chattopadhay, Anirban Sarkar, Prantik Howlader, and Vineeth~N
  Balasubramanian.
\newblock Grad-cam++: Generalized gradient-based visual explanations for deep
  convolutional networks.
\newblock In {\em WACV}, 2018.

\bibitem{this-that}
Chaofan Chen, Oscar Li, Chaofan Tao, Alina~Jade Barnett, Jonathan Su, and
  Cynthia Rudin.
\newblock This looks like that: Deep learning for interpretable image
  recognition.
\newblock In {\em NeurIPS}, 2019.

\bibitem{kd_1}
Pengguang Chen, Shu Liu, Hengshuang Zhao, and Jiaya Jia.
\newblock Distilling knowledge via knowledge review.
\newblock In {\em IEEE Conference on Computer Vision and Pattern Recognition
  (CVPR)}, 2021.

\bibitem{imagenet}
J. {Deng}, W. {Dong}, R. {Socher}, L. {Li}, {Kai Li}, and {Li Fei-Fei}.
\newblock Imagenet: A large-scale hierarchical image database.
\newblock In {\em CVPR}, 2009.

\bibitem{ablation-cam}
Saurabh Desai and Harish~G. Ramaswamy.
\newblock Ablation-cam: Visual explanations for deep convolutional network via
  gradient-free localization.
\newblock In {\em WACV}, 2020.

\bibitem{extremalPerturbations}
Ruth Fong, Mandela Patrick, and Andrea Vedaldi.
\newblock Understanding deep networks via extremal perturbations and smooth
  masks.
\newblock In {\em ICCV}, 2019.

\bibitem{xgrad-cam}
Ruigang Fu, Qingyong Hu, Xiaohu Dong, Yulan Guo, Yinghui Gao, and Biao Li.
\newblock Axiom-based grad-cam: Towards accurate visualization and explanation
  of cnns.
\newblock In {\em BMVC}, 2020.

\bibitem{scg}
Yunhao Ge, Yao Xiao, Zhi Xu, Meng Zheng, Srikrishna Karanam, Terrence Chen,
  Laurent Itti, and Ziyan Wu.
\newblock A peek into the reasoning of neural networks: Interpreting with
  structural visual concepts.
\newblock In {\em CVPR}, 2021.

\bibitem{glorot}
Xavier Glorot, Antoine Bordes, and Yoshua Bengio.
\newblock Deep sparse rectifier neural networks.
\newblock In {\em AISTATS}, volume~15, 2011.

\bibitem{Goodfellow-et-al-2016}
Ian Goodfellow, Yoshua Bengio, and Aaron Courville.
\newblock {\em Deep Learning}.
\newblock MIT Press, 2016.
\newblock \url{http://www.deeplearningbook.org}.

\bibitem{clrp}
Jindong Gu, Yinchong Yang, and Volker Tresp.
\newblock Understanding individual decisions of cnns via contrastive
  backpropagation.
\newblock In {\em ACCV}, 2019.

\bibitem{recursive-net}
Qiushan Guo, Zhipeng Yu, Yichao Wu, Ding Liang, Haoyu Qin, and Junjie Yan.
\newblock Dynamic recursive neural network.
\newblock In {\em CVPR}, 2019.

\bibitem{kd_2}
Geoffrey~E. Hinton, Oriol Vinyals, and Jeffrey Dean.
\newblock Distilling the knowledge in a neural network.
\newblock {\em ArXiv}, abs/1503.02531, 2015.

\bibitem{roar}
Sara Hooker, D. Erhan, Pieter-Jan Kindermans, and Been Kim.
\newblock A benchmark for interpretability methods in deep neural networks.
\newblock In {\em NeurIPS}, 2019.

\bibitem{global_attributions}
Mark Ibrahim, Melissa Louie, Ceena Modarres, and John Paisley.
\newblock Global explanations of neural networks: Mapping the landscape of
  predictions.
\newblock In {\em Proceedings of the 2019 AAAI/ACM Conference on AI, Ethics,
  and Society}, 2019.

\bibitem{PathwayGrad}
Ashkan Khakzar, Soroosh Baselizadeh, Saurabh Khanduja, Christian Rupprecht,
  Seong~Tae Kim, and Nassir Navab.
\newblock Neural response interpretation through the lens of critical pathways.
\newblock In {\em CVPR}, 2021.

\bibitem{daq}
D. Kim, J. Lee, and B. Ham.
\newblock Distance-aware quantization.
\newblock In {\em ICCV}, 2021.

\bibitem{xprotonet}
Eunji Kim, Siwon Kim, Minji Seo, and Sungroh Yoon.
\newblock Xprotonet: Diagnosis in chest radiography with global and local
  explanations.
\newblock In {\em CVPR}, 2021.

\bibitem{class_fir_1}
Gabriel Kreiman, Christof Koch, and Itzhak Fried.
\newblock Category-specific visual responses of single neurons in the human
  medial temporal lobe.
\newblock {\em Nature Neuroscience}, 3, Sep 2000.

\bibitem{cifar10}
Alex Krizhevsky.
\newblock Learning multiple layers of features from tiny images.
\newblock Technical report, 2009.

\bibitem{alexnet}
Alex Krizhevsky, Ilya Sutskever, and Geoffrey~E Hinton.
\newblock Imagenet classification with deep convolutional neural networks.
\newblock In {\em NeurIPS}, 2012.

\bibitem{lecun_pruning}
Yann LeCun, John Denker, and Sara Solla.
\newblock Optimal brain damage.
\newblock In {\em NeurIPS}, 1989.

\bibitem{relevance-cam}
Jeong~Ryong Lee, Sewon Kim, Inyong Park, Taejoon Eo, and Dosik Hwang.
\newblock Relevance-cam: Your model already knows where to look.
\newblock In {\em CVPR}, 2021.

\bibitem{lfi-cam}
Kwang~Hee Lee, Chaewon Park, Junghyun Oh, and Nojun Kwak.
\newblock Lfi-cam: Learning feature importance for better visual explanation.
\newblock In {\em ICCV}, 2021.

\bibitem{relex}
Dohun Lim, Hyeonseok Lee, and Sungchan Kim.
\newblock Building reliable explanations of unreliable neural networks: Locally
  smoothing perspective of model interpretation.
\newblock In {\em CVPR}, 2021.

\bibitem{lrp}
Grégoire Montavon, Sebastian Lapuschkin, Alexander Binder, Wojciech Samek, and
  Klaus-Robert Müller.
\newblock Explaining nonlinear classification decisions with deep taylor
  decomposition.
\newblock {\em Pattern Recognition}, 65, 2017.

\bibitem{pmlr-v70-nagamine17a}
Tasha Nagamine and Nima Mesgarani.
\newblock Understanding the representation and computation of multilayer
  perceptrons: A case study in speech recognition.
\newblock In {\em ICML}, 2017.

\bibitem{Natarajan1995SparseAS}
Balas~K. Natarajan.
\newblock Sparse approximate solutions to linear systems.
\newblock {\em SIAM J. Comput.}, 24, 1995.

\bibitem{proto-tree}
Meike Nauta, Ron van Bree, and Christin Seifert.
\newblock Neural prototype trees for interpretable fine-grained image
  recognition.
\newblock In {\em CVPR}, 2021.

\bibitem{ucam}
Badri Patro, Mayank Lunayach, Shivansh Patel, and Vinay Namboodiri.
\newblock U-cam: Visual explanation using uncertainty based class activation
  maps.
\newblock In {\em ICCV}, 2019.

\bibitem{rise}
Vitali Petsiuk, Abir Das, and Kate Saenko.
\newblock Rise: Randomized input sampling for explanation of black-box models.
\newblock In {\em BMVC}, 2018.

\bibitem{prob_pruning}
Xin-Yao Qian and Diego Klabjan.
\newblock A probabilistic approach to neural network pruning.
\newblock In {\em ICML}, 2021.

\bibitem{class_fir_2}
R.~Quian Quiroga, L. Reddy, G. Kreiman, C. Koch, and I. Fried.
\newblock Invariant visual representation by single neurons in the human brain.
\newblock {\em Nature}, 435, 2005.

\bibitem{Ras2018ExplanationMI}
Gabrielle Ras, Marcel van Gerven, and W.F.G~Pim Haselager.
\newblock Explanation methods in deep learning: Users, values, concerns and
  challenges.
\newblock {\em ArXiv}, abs/1803.07517, 2018.

\bibitem{class_fir_3}
Leila Reddy and Nancy Kanwisher.
\newblock Coding of visual objects in the ventral stream.
\newblock {\em Current Opinion in Neurobiology}, 16, 2006.

\bibitem{LIME}
Marco~Tulio Ribeiro, Sameer Singh, and Carlos Guestrin.
\newblock "why should i trust you?": Explaining the predictions of any
  classifier.
\newblock In {\em SIGKDD}, 2016.

\bibitem{grad-cam}
Ramprasaath~R Selvaraju, Michael Cogswell, Abhishek Das, Ramakrishna Vedantam,
  Devi Parikh, and Dhruv Batra.
\newblock Grad-cam: Visual explanations from deep networks via gradient-based
  localization.
\newblock In {\em ICCV}, 2017.

\bibitem{deeplift}
Avanti Shrikumar, Peyton Greenside, and Anshul Kundaje.
\newblock Learning important features through propagating activation
  differences.
\newblock In {\em ICML}, 2017.

\bibitem{gradient}
Karen Simonyan, Andrea Vedaldi, and Andrew Zisserman.
\newblock Deep inside convolutional networks: Visualising image classification
  models and saliency maps.
\newblock In {\em ICLR Workshop}, 2014.

\bibitem{vgg}
Karen Simonyan and Andrew Zisserman.
\newblock Very deep convolutional networks for large-scale image recognition.
\newblock In {\em ICLR}, 2015.

\bibitem{smoothGrad}
Daniel {Smilkov}, Nikhil {Thorat}, Been {Kim}, Fernanda {Vi{\'e}gas}, and
  Martin {Wattenberg}.
\newblock {SmoothGrad: removing noise by adding noise}.
\newblock In {\em ICML Workshop}, 2017.

\bibitem{guided-backprop}
J.T. Springenberg, A. Dosovitskiy, T. Brox, and M. Riedmiller.
\newblock Striving for simplicity: The all convolutional net.
\newblock In {\em ICLR Workshop}, 2015.

\bibitem{nesy-xil}
Wolfgang Stammer, Patrick Schramowski, and Kristian Kersting.
\newblock Right for the right concept: Revising neuro-symbolic concepts by
  interacting with their explanations.
\newblock In {\em CVPR}, 2021.

\bibitem{integrated}
Mukund Sundararajan, Ankur Taly, and Qiqi Yan.
\newblock Axiomatic attribution for deep networks.
\newblock In {\em ICML}, 2017.

\bibitem{tsne}
Laurens van~der Maaten and Geoffrey Hinton.
\newblock Visualizing data using t-sne.
\newblock {\em Journal of Machine Learning Research}, 9, 2008.

\bibitem{FGVis}
Jorg Wagner, Jan~Mathias Kohler, Tobias Gindele, Leon Hetzel, Jakob~Thaddaus
  Wiedemer, and Sven Behnke.
\newblock Interpretable and fine-grained visual explanations for convolutional
  neural networks.
\newblock In {\em CVPR}, 2019.

\bibitem{score-cam}
Haofan Wang, Mengnan Du, Fan Yang, and Zijian Zhang.
\newblock Score-cam: Improved visual explanations via score-weighted class
  activation mapping.
\newblock In {\em CVPRW}, 2020.

\bibitem{wang2021neural}
Huan Wang, Can Qin, Yulun Zhang, and Yun Fu.
\newblock Neural pruning via growing regularization.
\newblock In {\em ICLR}, 2021.

\bibitem{pmlr-v139-wang21e}
Wenxiao Wang, Minghao Chen, Shuai Zhao, Long Chen, Jinming Hu, Haifeng Liu,
  Deng Cai, Xiaofei He, and Wei Liu.
\newblock Accelerate cnns from three dimensions: A comprehensive pruning
  framework.
\newblock In {\em ICLR}, 2021.

\bibitem{dgr}
Yulong Wang, Hang Su, Bo Zhang, and Xiaolin Hu.
\newblock Interpret neural networks by identifying critical data routing paths.
\newblock In {\em CVPR}, 2018.

\bibitem{cr-distillation}
Fuxun Yu, Zhuwei Qin, and Xiang Chen.
\newblock Distilling critical paths in convolutional neural networks.
\newblock In {\em NeurIPSW}, 2018.

\bibitem{deconvnet}
Matthew~D. Zeiler and Rob Fergus.
\newblock Visualizing and understanding convolutional networks.
\newblock In {\em ECCV}, 2014.

\bibitem{excitation}
Jianming Zhang, Zhe Lin, Jonathan Brandt, Xiaohui Shen, and Stan Sclaroff.
\newblock Top-down neural attention by excitation backprop.
\newblock {\em IJCV}, 126, 2017.

\bibitem{cam}
Bolei Zhou, Aditya Khosla, Agata Lapedriza, Aude Oliva, and Antonio Torralba.
\newblock Learning deep features for discriminative localization.
\newblock In {\em CVPR}, 2016.

\end{thebibliography}
}

\clearpage

\section{Supplementary Materials}
\subsection{Experimental Details}
\paragraphX{Model Training.} We use the code repository \footnote{https://github.com/bearpaw/pytorch-classification} to train the models on CIFAR-10 \cite{cifar10} and ImageNet \cite{imagenet}. The training set and test set splits of CIFAR-10 follow the original data split \cite{cifar10}. To train the ImageNet models, we train on the validation set and rescale the image size to 32 to minimize computation time and due to limited computation resources. All our trainings and tests are performed on a NVIDIA RTX-3090. The hyperparameters used for our models are in Table \ref{tab:gen-cnp-train}.

\begin{table}[!htp]\centering
\caption{Hyperparameters for GEN-CNP}\label{tab:gen-cnp-train}
\scriptsize
\begin{tabular}{l| c | c | c | c}
\toprule
Dataset &Model &alpha &beta &learning rate \\
\midrule
\multirow{2}{*}{CIFAR-10} &AlexNet &1 &0.05 &0.0001 \\
&VGG-11 &1 &0.0005 &0.09 \\
\multirow{2}{*}{ImageNet} &AlexNet &1 &0.005 &0.001 \\
&VGG-11 &1 &0.005 &0.005 \\
\bottomrule
\end{tabular}
\end{table}

\subsection{Additional Quantitative Results}

Another commonly used metric is the Increase in Confidence Rate (ICr). Originally, ICr in \cite{grad-cam++} measures the average number of samples with increased confidence regardless of the classification accuracy as $\sum_i^n \text{sign} (\hat{y}_i > y_i) / n$ where $\hat{y}_i$ is the model's probability on the neural pathways, $y$ is the model's original probability and $n$ is the total number of samples in the data set. We instead propose to consider only the correctly classified samples, which is much stricter, thus $\text{ICr} = \sum_i^n \mathds{1}_{ \{(\hat{y}_i > y_i) \wedge \hat{c}_i = c_i\} } / n$ where $\hat{c}_i = \argmax (\hat{y}_i)$ and $c_i = \argmax (y_i)$.
    
In Table 1 of the paper, we report the results of the quantitative results on ImageNet. In Table \ref{tab:cifar10-fait}, we report the results of the quantitaive results on CIFAR-10 \cite{cifar10}. We observe similarly that our model GEN-CNP is able to achieve better performance than the baselines on the faithfulness benchmarks. 

\begin{table*}[!htp]\centering
\scriptsize
\setlength{\extrarowheight}{0.5mm}
\caption{Results (\%) on AlexNet \cite{alexnet} and VGG-11 \cite{vgg} models for CIFAR-10 \cite{cifar10}. The test accuracies of AlexNet and VGG-11 are $77.46\%$ and $92.56\%$. The Accuracy metric is for reference only.}
\begin{tabular}{l|cccc|cccc}
\hline
\multirow{2}{*}{Method} & \multicolumn{4}{c|}{AlexNet} & \multicolumn{4}{c}{VGG11} \\
\cline{2-9}
 & ICr ($\uparrow$) & mIC ($\uparrow$) & mDC ($\downarrow$) & Accuracy ($\uparrow$) & ICr ($\uparrow$) & mIC ($\uparrow$) & mDC ($\downarrow$) & Accuracy ($\uparrow$) \\
\hline
Greedy \cite{PathwayGrad} &6.31 &0.80  &24.59  &38.87  &1.03  &9.54  &26.37  &25.11  \\
Random &0.00 &0.00  &67.61  &10.08  &0.00  &0.00  &81.93  &9.78  \\
Original Activation \cite{lecun_pruning} &1.30  &6.76  &35.84  &38.70  &1.12 &6.79  &26.01 &92.13  \\
\hline
NeuronMCT \cite{PathwayGrad} &0.02 &5.37 &55.30 &77.22 &0.01 &2.64 &64.24  &92.35  \\
NeuronGrad \cite{PathwayGrad} &0.02 &6.17 &54.70 & 77.28 &0.01 &1.51 &59.08 &92.41 \\
DGR \cite{dgr} &1.49 &12.82 &47.98 &76.79 &0.61 &5.81 &24.84 & 92.44 \\
DGR\_R \cite{dgr} &0.01 &5.79 &28.79 &77.11 &0.00 &0.00 &57.34 &92.43 \\
\textbf{GEN-CNP (ours)} &\textbf{13.20} &\textbf{14.67} &\textbf{14.40} &77.12 &\textbf{4.01} &\textbf{10.26} &\textbf{9.04} &92.14 \\
\hline
\end{tabular}
\label{tab:cifar10-fait}
\end{table*}

\section{acIOU}
\begin{figure*}[h]
\begin{subfigure}{.3\textwidth}
  \centering
  \includegraphics[width=1\linewidth]{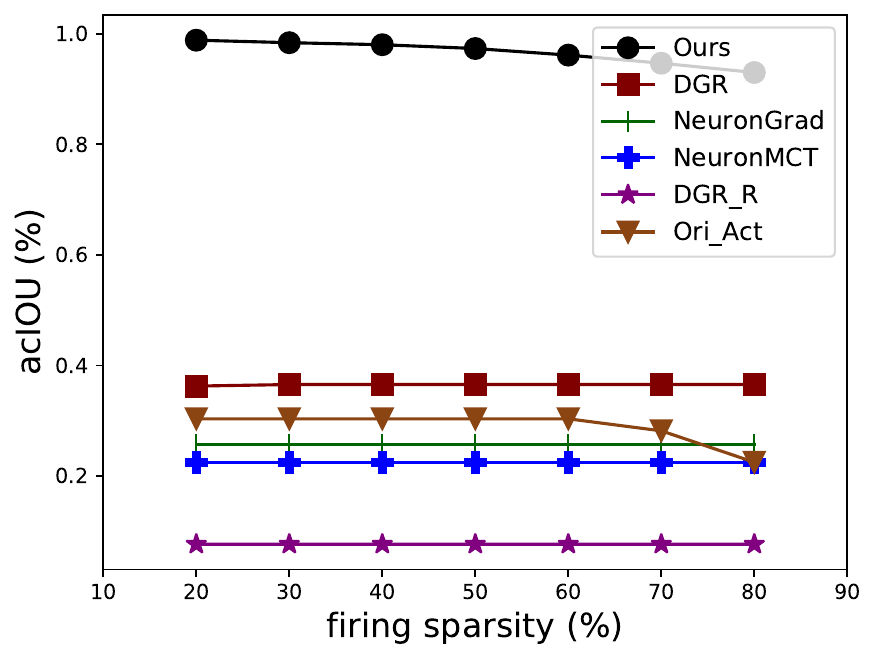}
  \caption{acIOU vs firing sparsity on VGG-11 \& CIFAR-10}
  \label{fig:ciou_ps_vgg_cifar}
\end{subfigure}
\hspace{0.3cm}
\begin{subfigure}{.3\textwidth}
  \centering
  \includegraphics[width=1\linewidth]{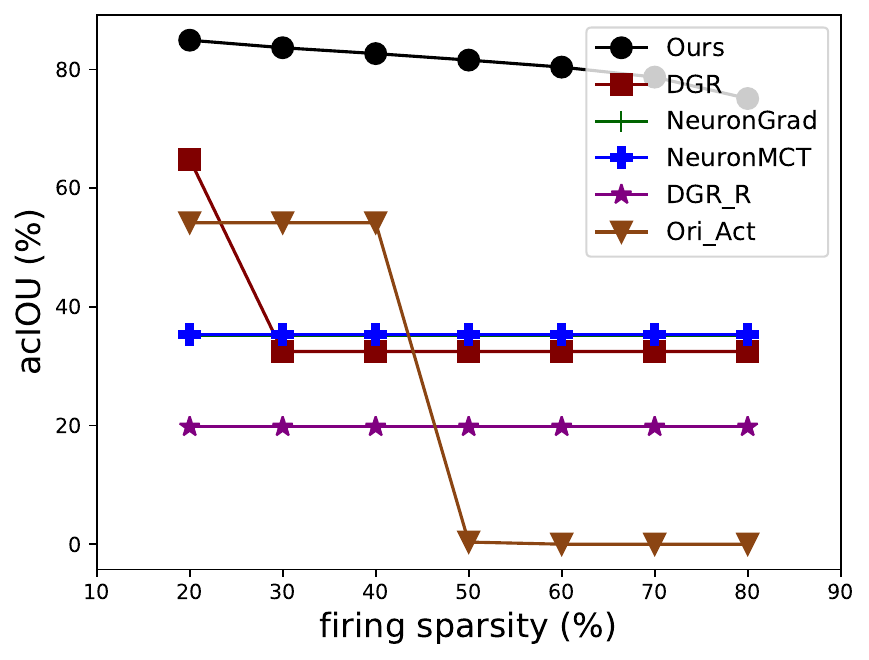}
  \caption{acIOU vs firing sparsity on VGG-11 \& ImageNet}
  \label{fig:ciou_ps_vgg_image}
\end{subfigure} 
\hspace{0.3cm}
\begin{subfigure}{.3\textwidth}
  \centering
  \includegraphics[width=1\linewidth]{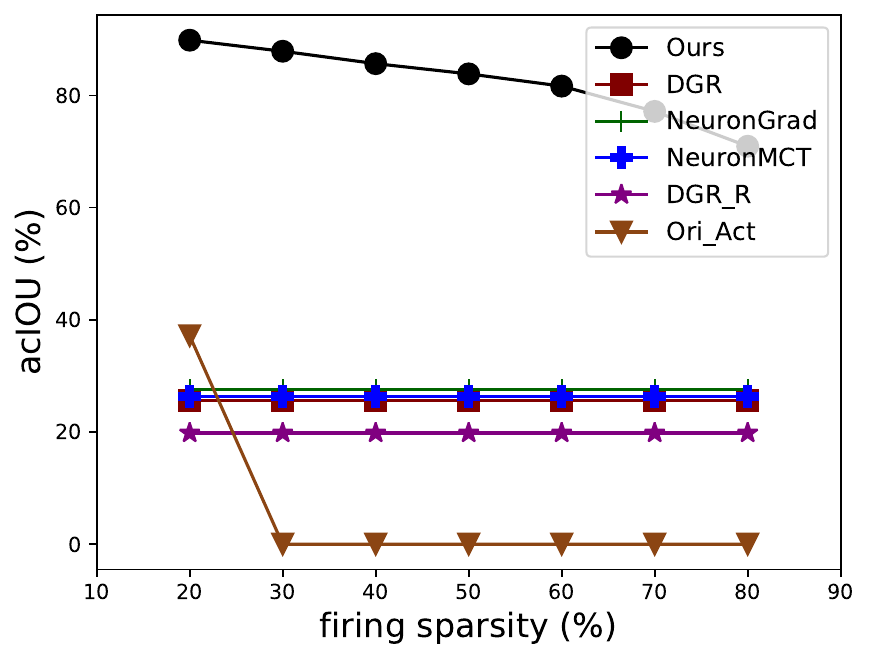}
  \caption{acIOU vs firing sparsity on AlexNet \& ImageNet}
  \label{fig:ciou_ps_alexnet_image}
\end{subfigure} \\
\begin{subfigure}{.3\textwidth}
  \centering
  \includegraphics[width=1\linewidth]{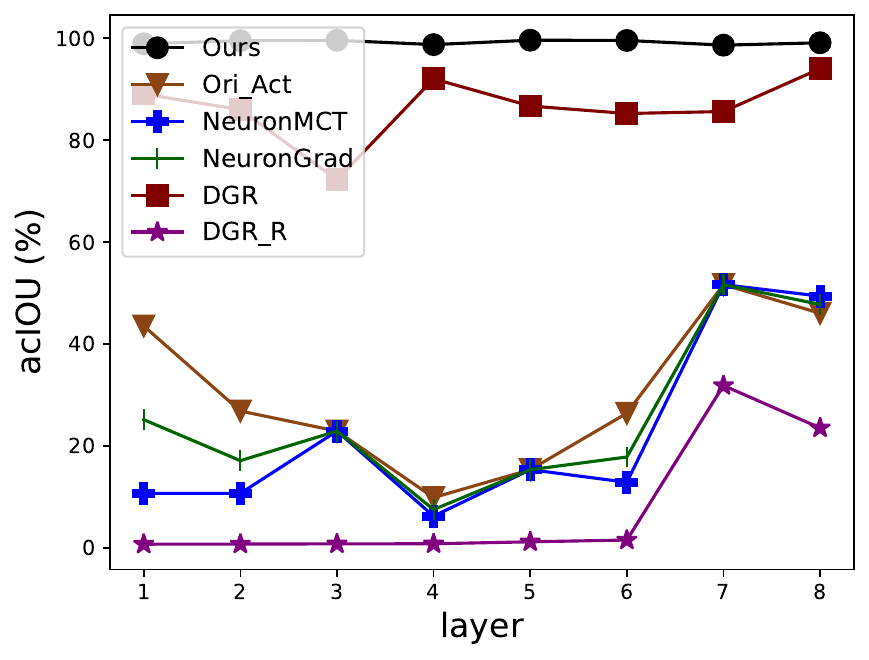}
  \caption{acIOU vs layer on VGG-11 \& CIFAR-10}
  \label{fig:ciou_layer_vgg_cifar}
\end{subfigure}%
\hspace{0.4cm}
\begin{subfigure}{.3\textwidth}
  \centering
  \includegraphics[width=1\linewidth]{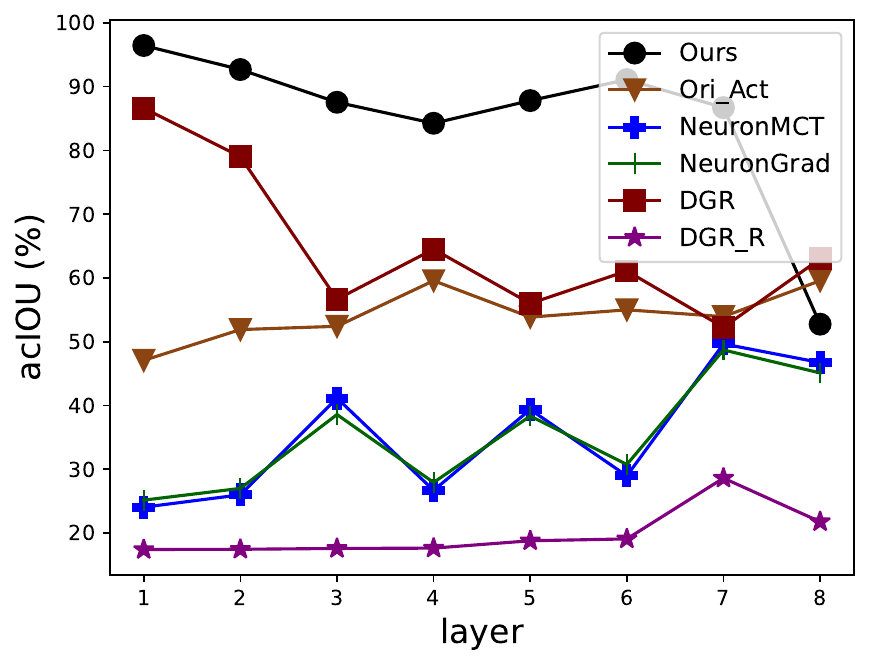}
  \caption{acIOU vs layer on VGG-11 \& ImageNet}
  \label{fig:ciou_layer_vgg_image}
\end{subfigure}%
\hspace{0.4cm}
\begin{subfigure}{.3\textwidth}
  \centering
  \includegraphics[width=1\linewidth]{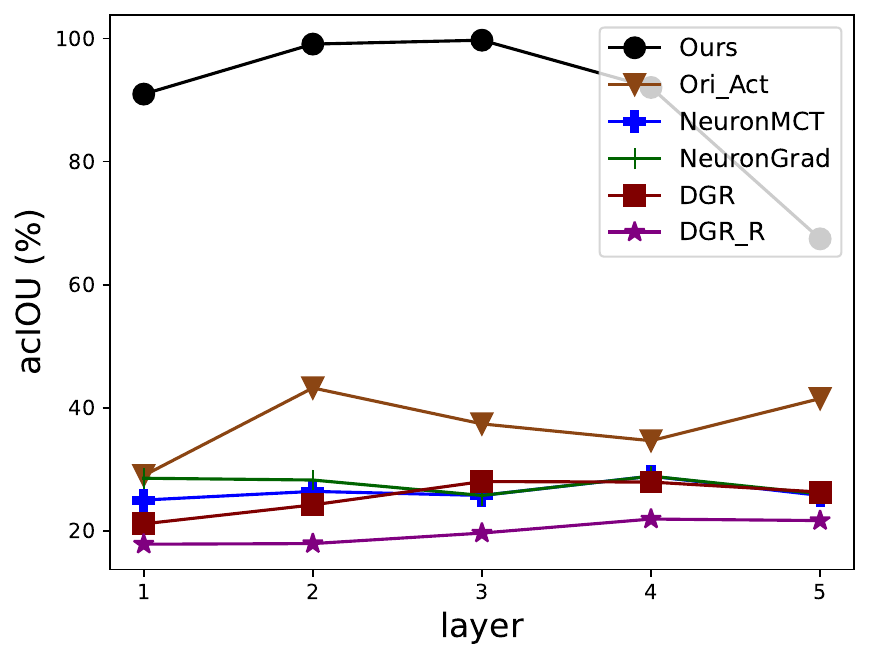}
  \caption{acIOU vs layer on AlexNet \& ImageNet}
  \label{fig:ciou_layer_alexnet_image}
\end{subfigure}%
\caption{\textbf{Additional Results on acIOU.} The first row shows the acIOU vs firing sparsity, the second row shows the acIOU vs layer for each model.}
\label{fig:aciou_supp}
\end{figure*}

We show in Fig. \ref{fig:aciou_supp} the results of acIOU on the model VGG-11 for CIFAR-10 and AlexNet and VGG-11 for ImageNet. These figures are to complement the results in the original paper Fig. 5(b) and 5(c). Our method GEN-CNP generates neural pathways with higher acIOU than all other methods, as shown in Fig. \ref{fig:ciou_ps_vgg_cifar}, \ref{fig:ciou_ps_vgg_image} and \ref{fig:ciou_ps_alexnet_image}. Fig. \ref{fig:ciou_layer_vgg_cifar}, \ref{fig:ciou_layer_vgg_image} and \ref{fig:ciou_layer_alexnet_image} show the acIOU vs layers for different models on different datasets. 

\section{Transferability Experiments}
\begin{figure*}[h]
\begin{subfigure}{.24\textwidth}
  \centering
  \includegraphics[width=1.1\linewidth]{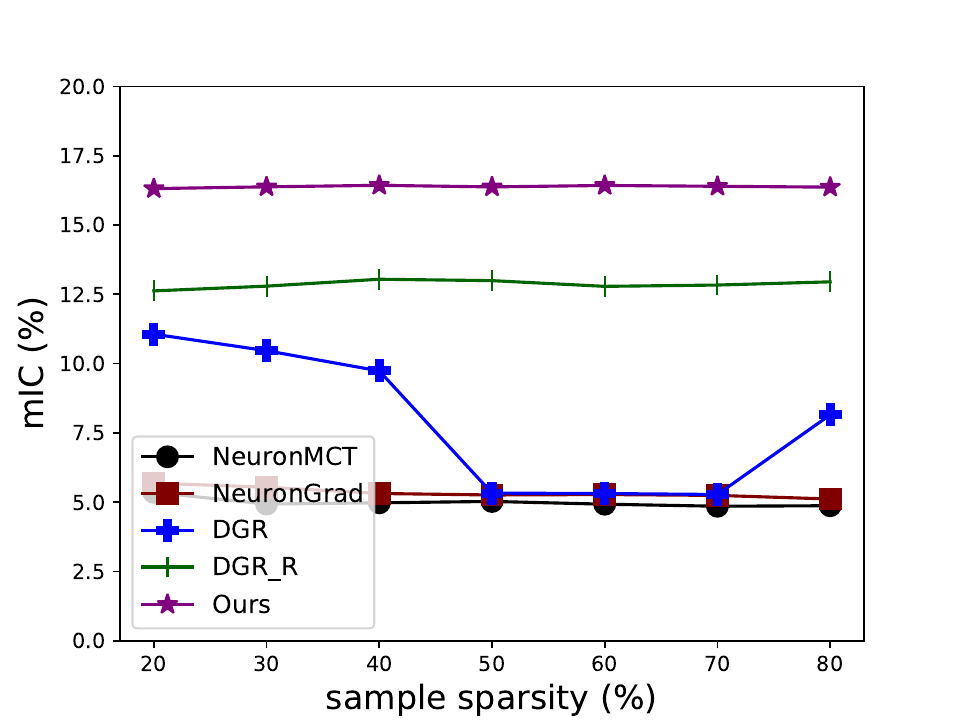}
  \caption{mIC vs $\epsilon_{ss}$, $\epsilon_{cn} = 0.1$}
  \label{fig:aic_vs_ms}
\end{subfigure}
\begin{subfigure}{.24\textwidth}
  \centering
  \includegraphics[width=1.1\linewidth]{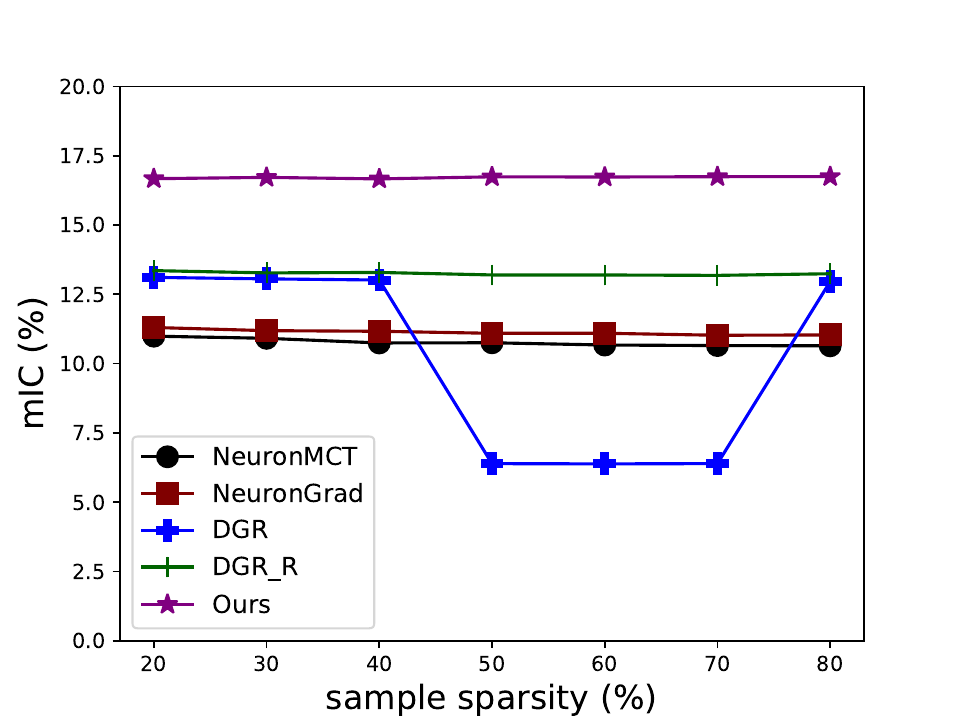}
  \caption{mIC vs $\epsilon_{ss}$, $\epsilon_{cn} = 0.2$}
  \label{fig:aic_vs_ms}
\end{subfigure}
\begin{subfigure}{.24\textwidth}
  \centering
  \includegraphics[width=1.1\linewidth]{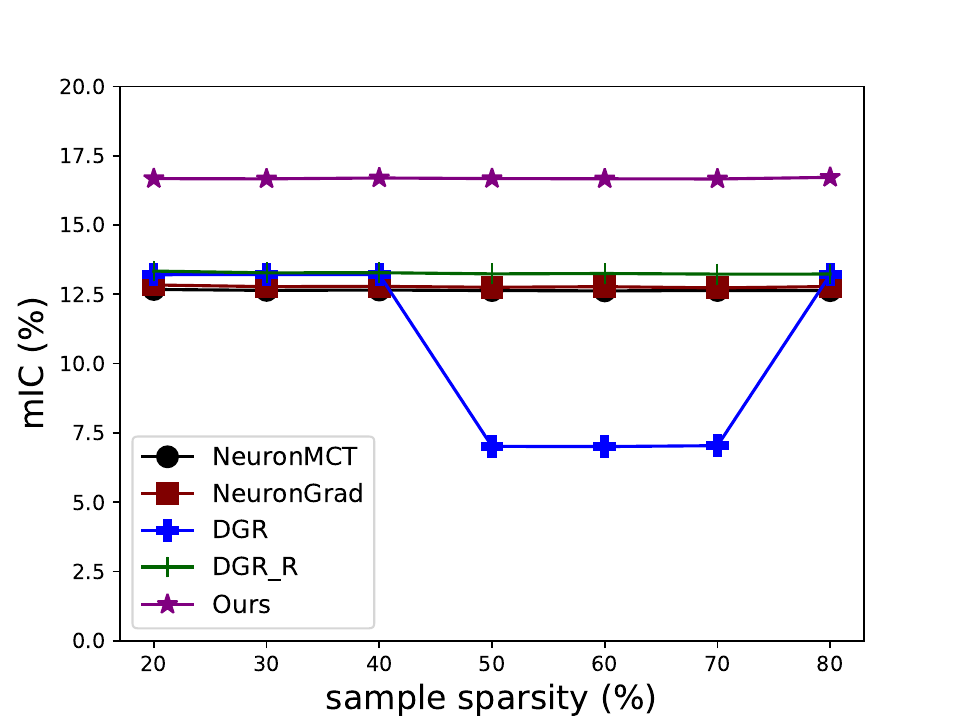}
  \caption{mIC vs $\epsilon_{ss}$, $\epsilon_{cn} = 0.3$}
  \label{fig:aic_vs_ms}
\end{subfigure}
\begin{subfigure}{.24\textwidth}
  \centering
  \includegraphics[width=1.1\linewidth]{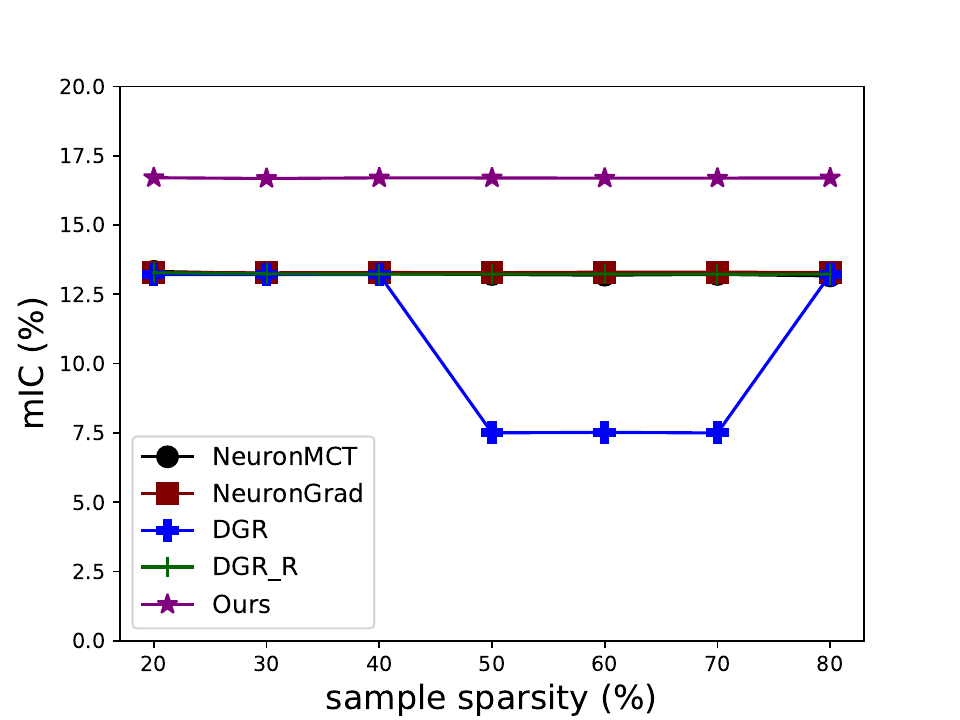}
  \caption{mIC vs $\epsilon_{ss}$, $\epsilon_{cn} = 0.4$}
  \label{fig:aic_vs_ms}
\end{subfigure}
\\
\begin{subfigure}{.24\textwidth}
  \centering
  \includegraphics[width=1.1\linewidth]{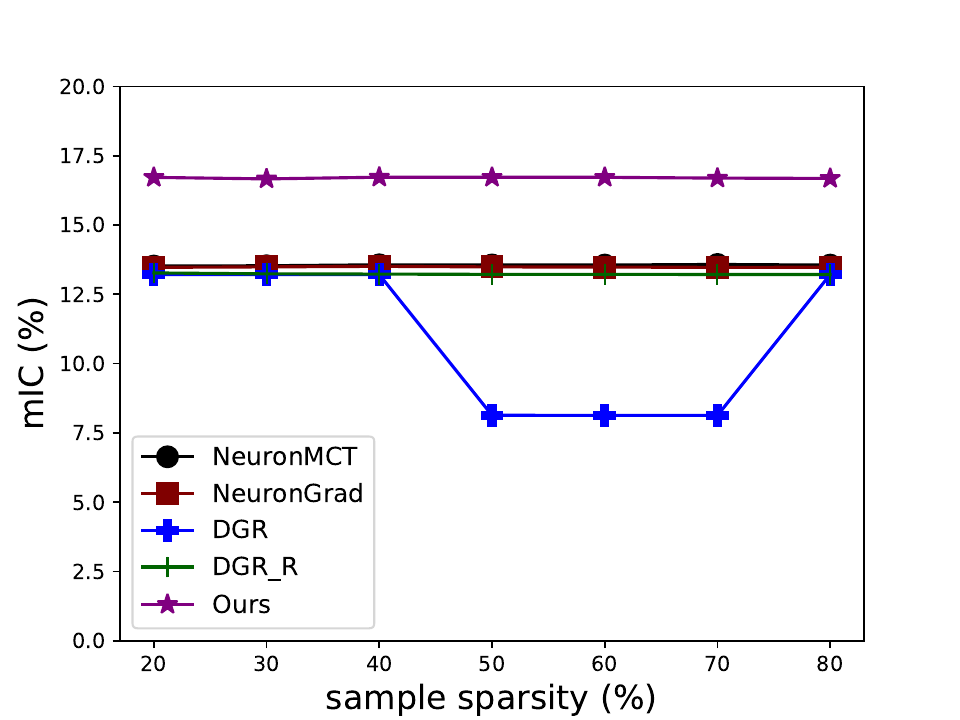}
  \caption{mIC vs $\epsilon_{ss}$, $\epsilon_{cn} = 0.5$}
  \label{fig:aic_vs_ms}
\end{subfigure} 
\begin{subfigure}{.24\textwidth}
  \centering
  \includegraphics[width=1.1\linewidth]{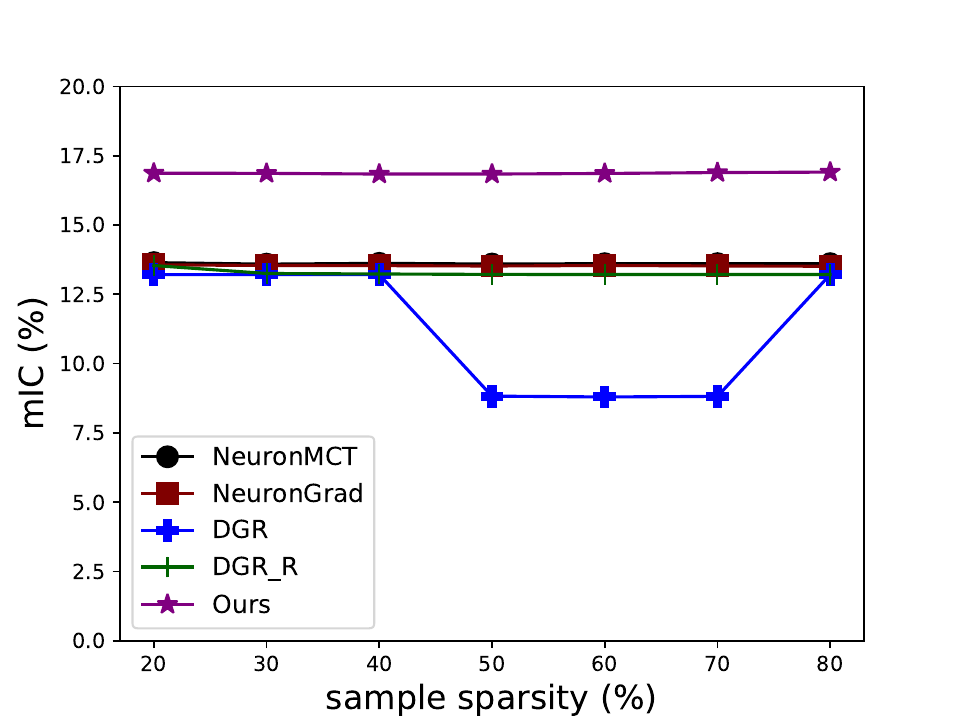}
  \caption{mIC vs $\epsilon_{ss}$, $\epsilon_{cn} = 0.6$}
  \label{fig:aic_vs_ms}
\end{subfigure}
\begin{subfigure}{.24\textwidth}
  \centering
  \includegraphics[width=1.1\linewidth]{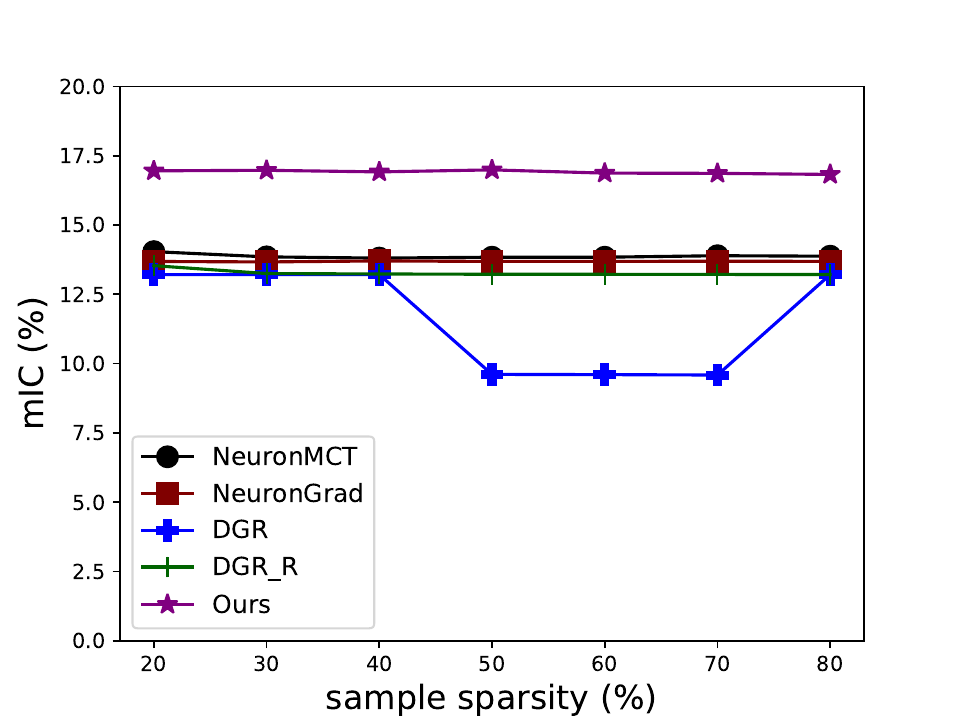}
  \caption{mIC vs $\epsilon_{ss}$, $\epsilon_{cn} = 0.7$}
  \label{fig:aic_vs_ms}
\end{subfigure}
\begin{subfigure}{.24\textwidth}
  \centering
  \includegraphics[width=1.1\linewidth]{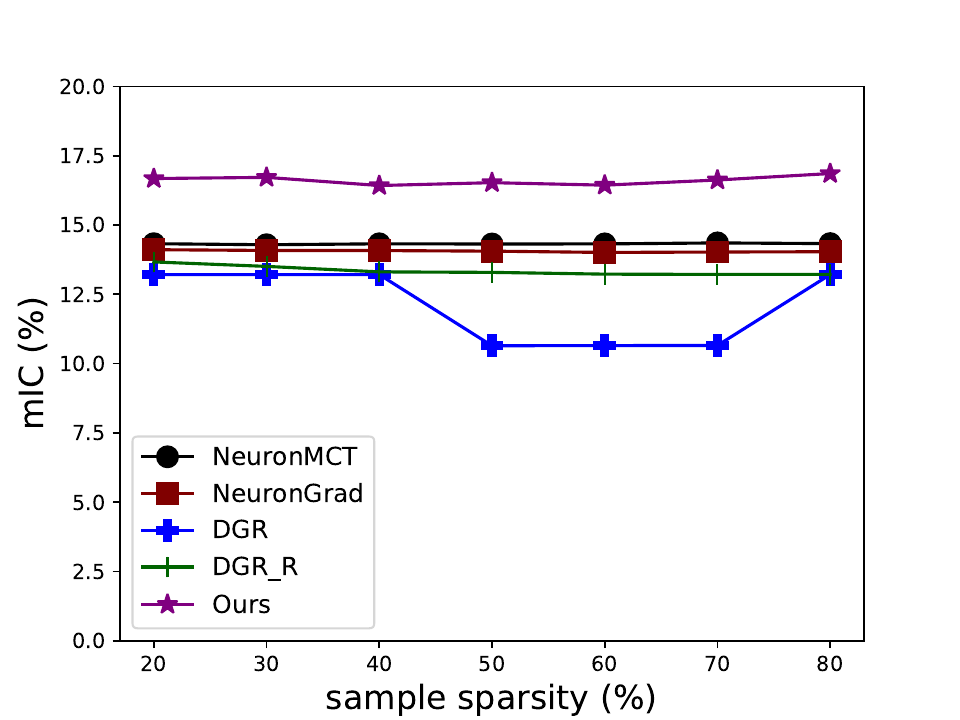}
  \caption{mIC vs $\epsilon_{ss}$, $\epsilon_{cn} = 0.8$}
  \label{fig:aic_vs_ms}
\end{subfigure}
\caption{\textbf{Transferability Experiments on VGG-11 \cite{vgg} \& CIFAR-10 \cite{cifar10}.} mIC vs samples sparsity for increasing class-relevant neural pathways firing sparsity $\epsilon_{cn}$.}
\label{fig:transfer_mic_vs_ss_vgg_cifar10}
\end{figure*}

\begin{figure*}[h]
\begin{subfigure}{.24\textwidth}
  \centering
  \includegraphics[width=1.1\linewidth]{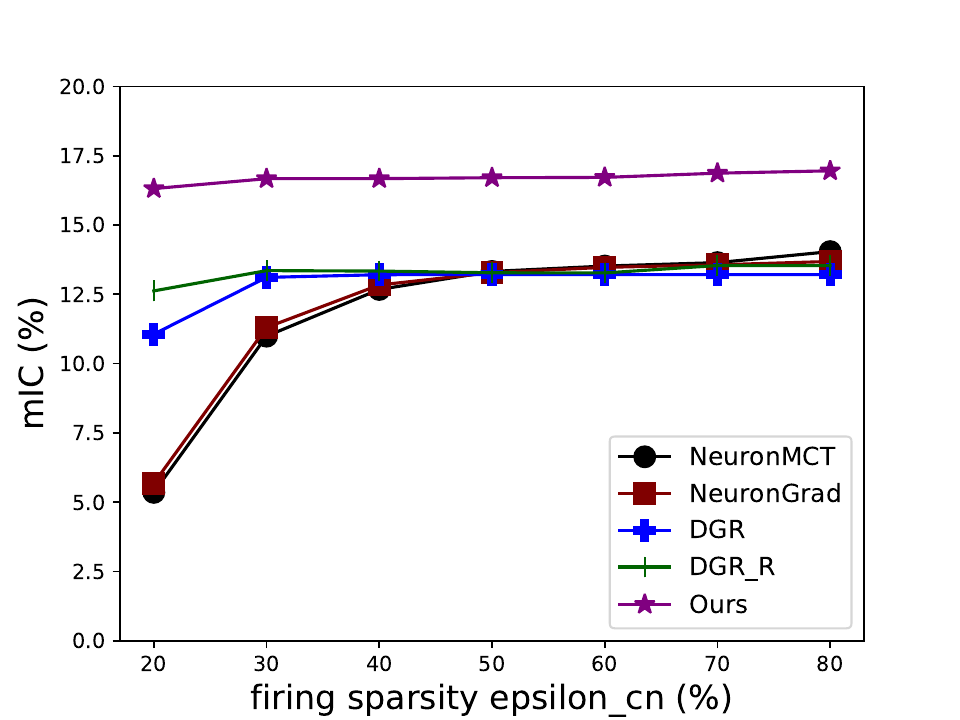}
  \caption{mIC vs $\epsilon_{cn}$, $\epsilon_{ss} = 0.2$}
  \label{fig:aic_vs_ms}
\end{subfigure}
\begin{subfigure}{.24\textwidth}
  \centering
  \includegraphics[width=1.1\linewidth]{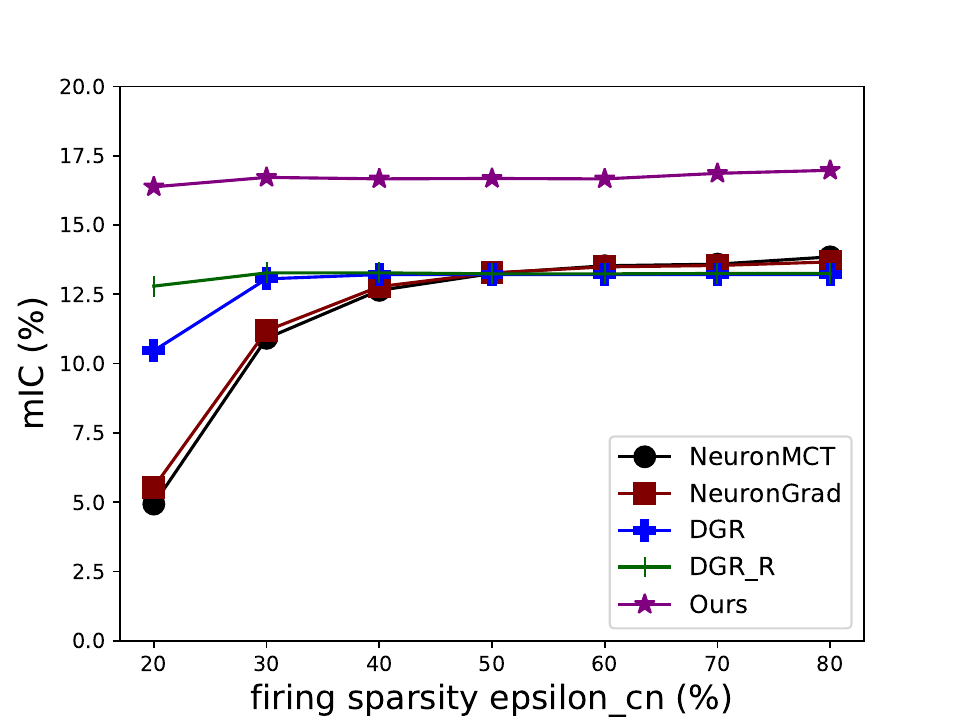}
  \caption{mIC vs $\epsilon_{cn}$, $\epsilon_{ss} = 0.3$}
  \label{fig:aic_vs_ms}
\end{subfigure}
\begin{subfigure}{.24\textwidth}
  \centering
  \includegraphics[width=1.1\linewidth]{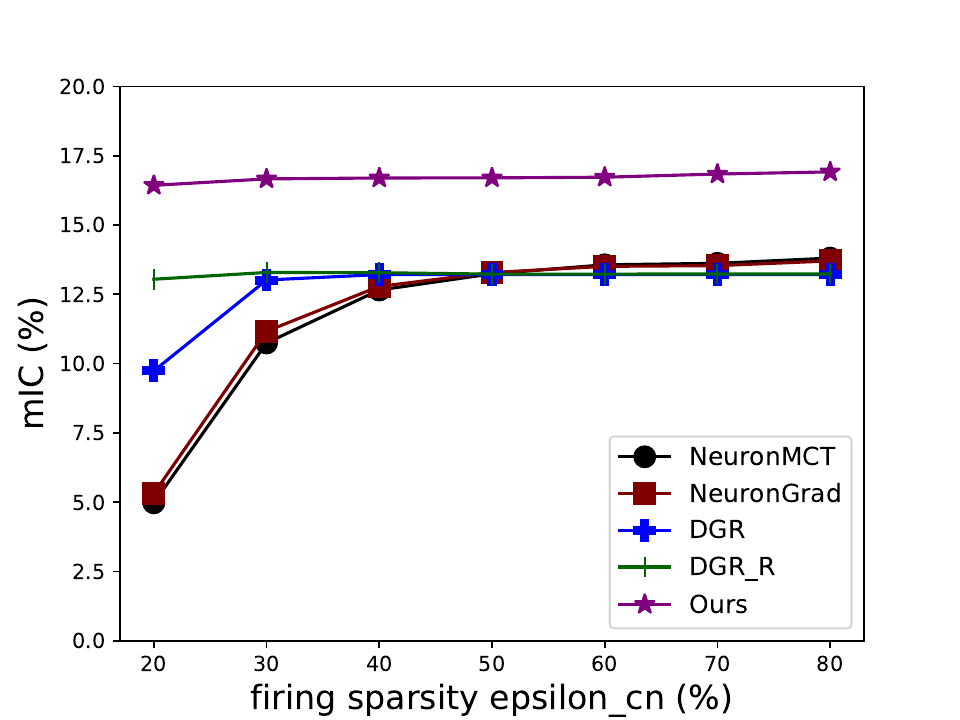}
  \caption{mIC vs $\epsilon_{cn}$, $\epsilon_{ss} = 0.4$}
  \label{fig:aic_vs_ms}
\end{subfigure}
\begin{subfigure}{.24\textwidth}
  \centering
  \includegraphics[width=1.1\linewidth]{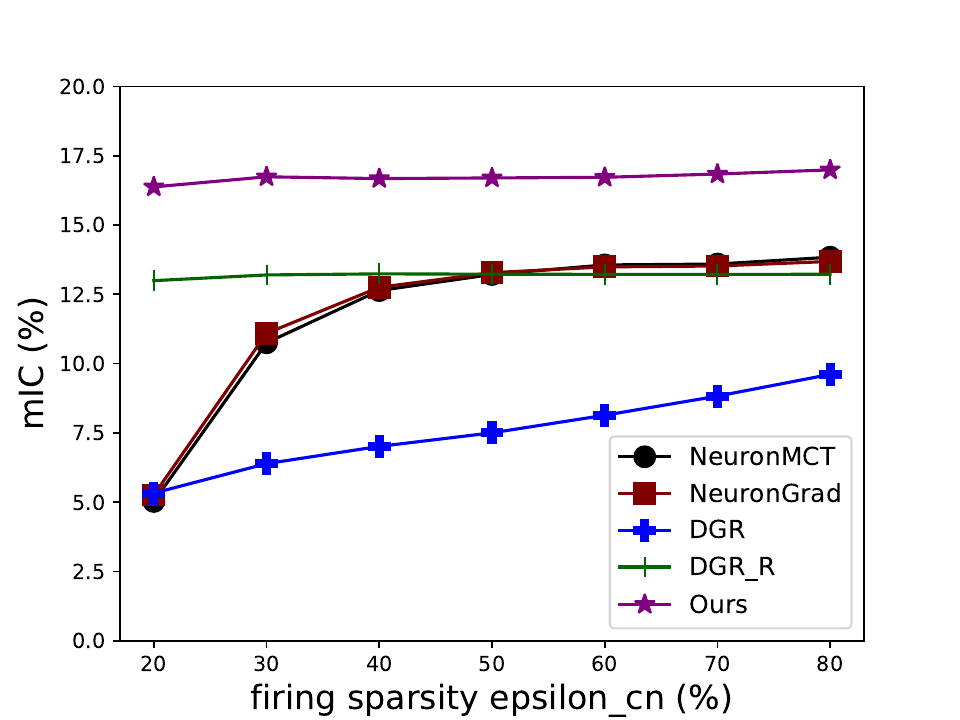}
  \caption{mIC vs $\epsilon_{cn}$, $\epsilon_{ss} = 0.5$}
  \label{fig:aic_vs_ms}
\end{subfigure}
\\
\begin{subfigure}{.24\textwidth}
  \centering
  \includegraphics[width=1.1\linewidth]{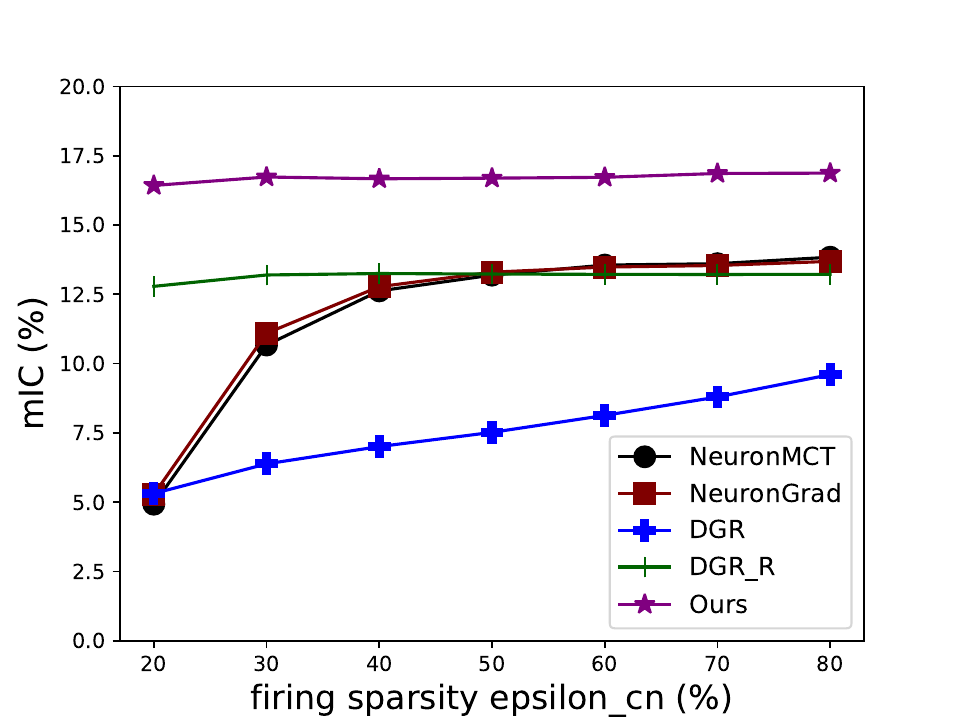}
  \caption{mIC vs $\epsilon_{cn}$, $\epsilon_{ss} = 0.6$}
  \label{fig:aic_vs_ms}
\end{subfigure} 
\begin{subfigure}{.24\textwidth}
  \centering
  \includegraphics[width=1.1\linewidth]{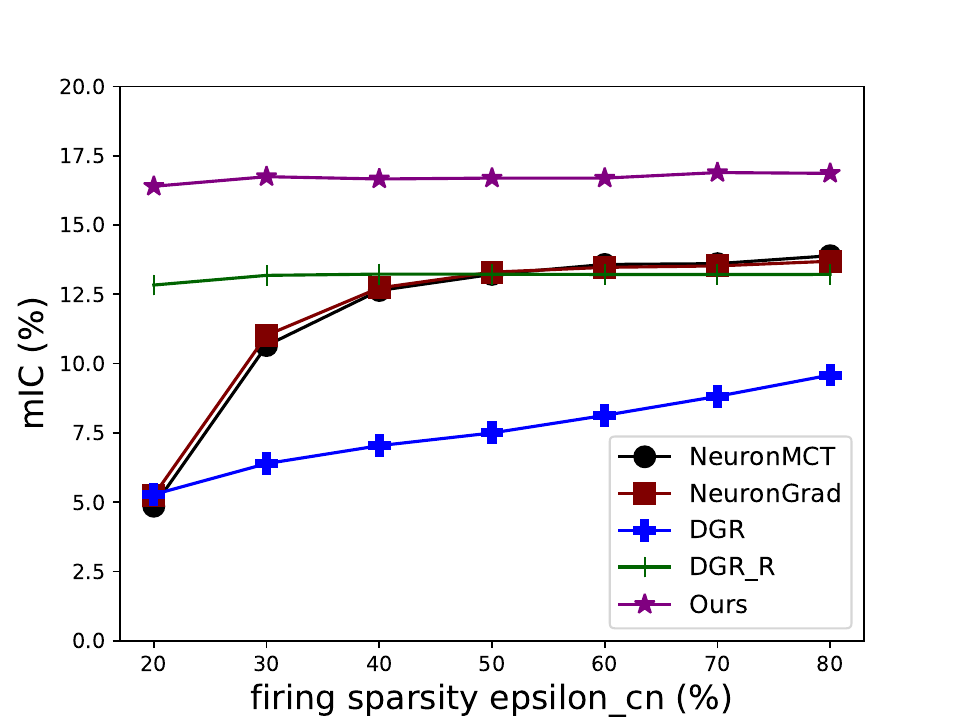}
  \caption{mIC vs $\epsilon_{cn}$, $\epsilon_{ss} = 0.7$}
  \label{fig:aic_vs_ms}
\end{subfigure}
\begin{subfigure}{.24\textwidth}
  \centering
  \includegraphics[width=1.1\linewidth]{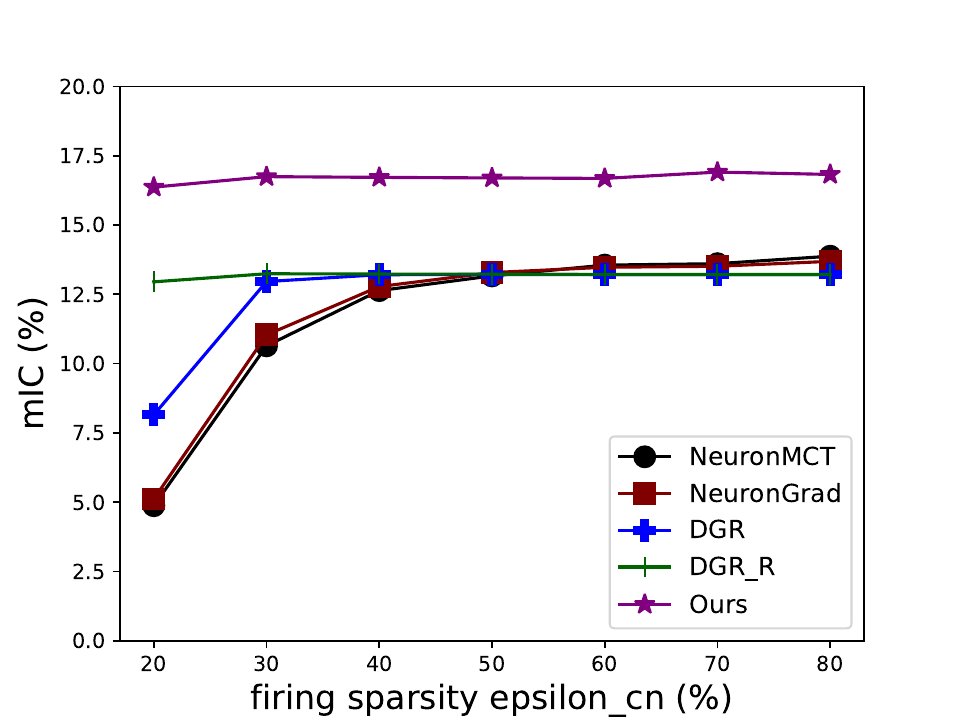}
  \caption{mIC vs $\epsilon_{cn}$, $\epsilon_{ss} = 0.8$}
  \label{fig:aic_vs_ms}
\end{subfigure}
\caption{\textbf{Transferability Experiments on VGG-11 \cite{vgg} \& CIFAR-10 \cite{cifar10}.} mIC vs class-relevant neural pathways firing sparsity $\epsilon_{cn}$ for different sample sparsity $\epsilon_{ss}$.}
\label{fig:transfer_alexnet_image}
\end{figure*}

\begin{figure*}[h]
\begin{subfigure}{.24\textwidth}
  \centering
  \includegraphics[width=1.1\linewidth]{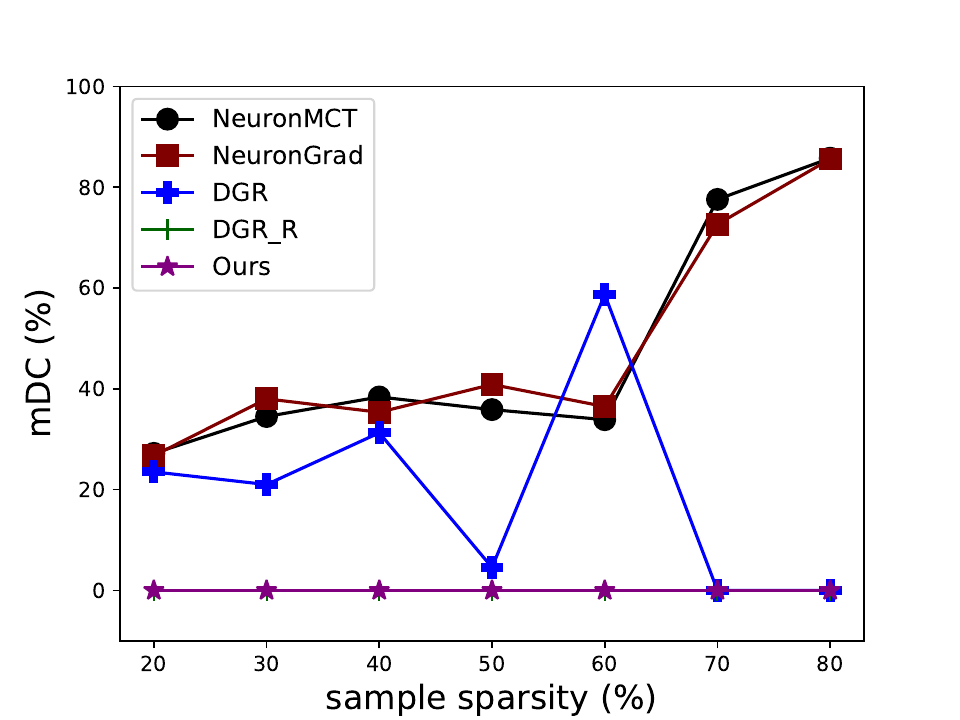}
  \caption{mDC vs $\epsilon_{ss}$, $\epsilon_{cn} = 0.2$}
  \label{fig:aic_vs_ms}
\end{subfigure}
\begin{subfigure}{.24\textwidth}
  \centering
  \includegraphics[width=1.1\linewidth]{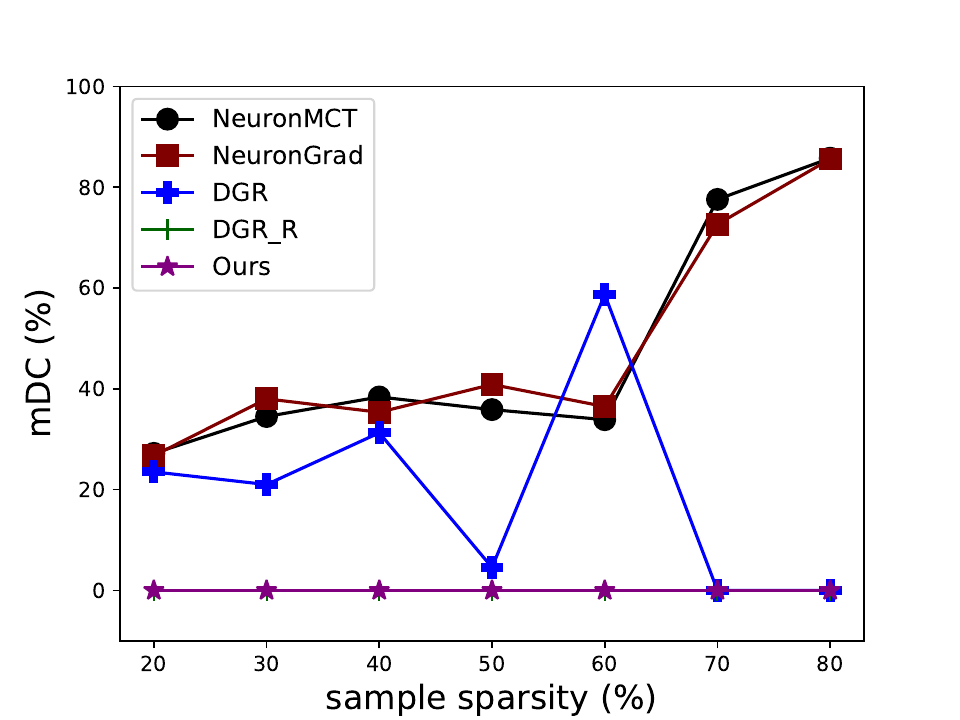}
  \caption{mDC vs $\epsilon_{ss}$, $\epsilon_{cn} = 0.3$}
  \label{fig:aic_vs_ms}
\end{subfigure}
\begin{subfigure}{.24\textwidth}
  \centering
  \includegraphics[width=1.1\linewidth]{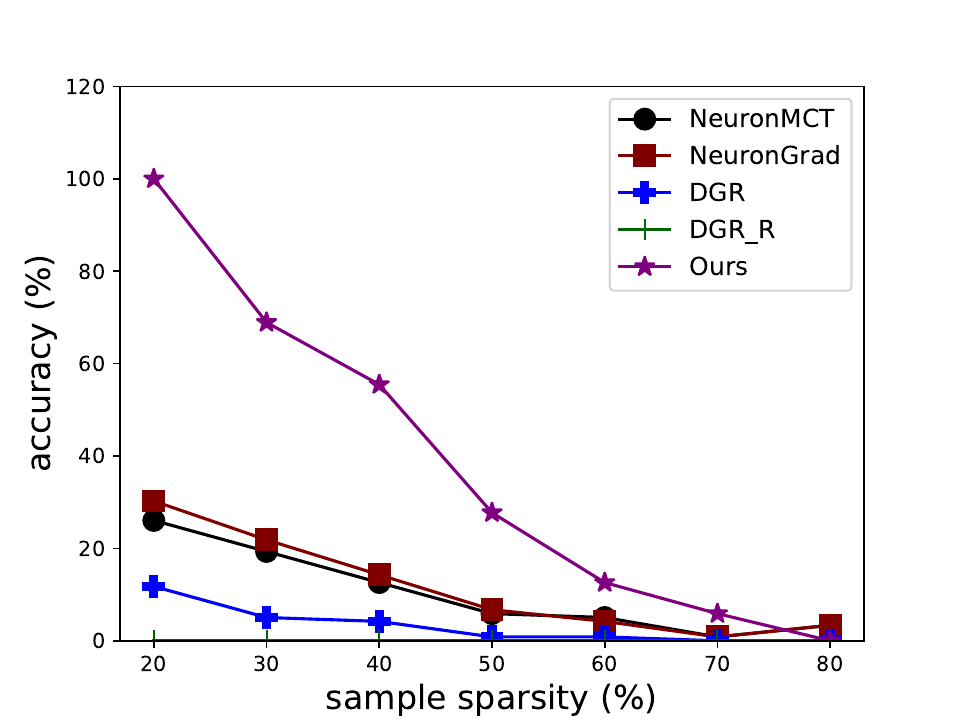}
  \caption{accuracy vs $\epsilon_{ss}$, $\epsilon_{cn} = 0.2$}
  \label{fig:aic_vs_ms}
\end{subfigure}
\begin{subfigure}{.24\textwidth}
  \centering
  \includegraphics[width=1.1\linewidth]{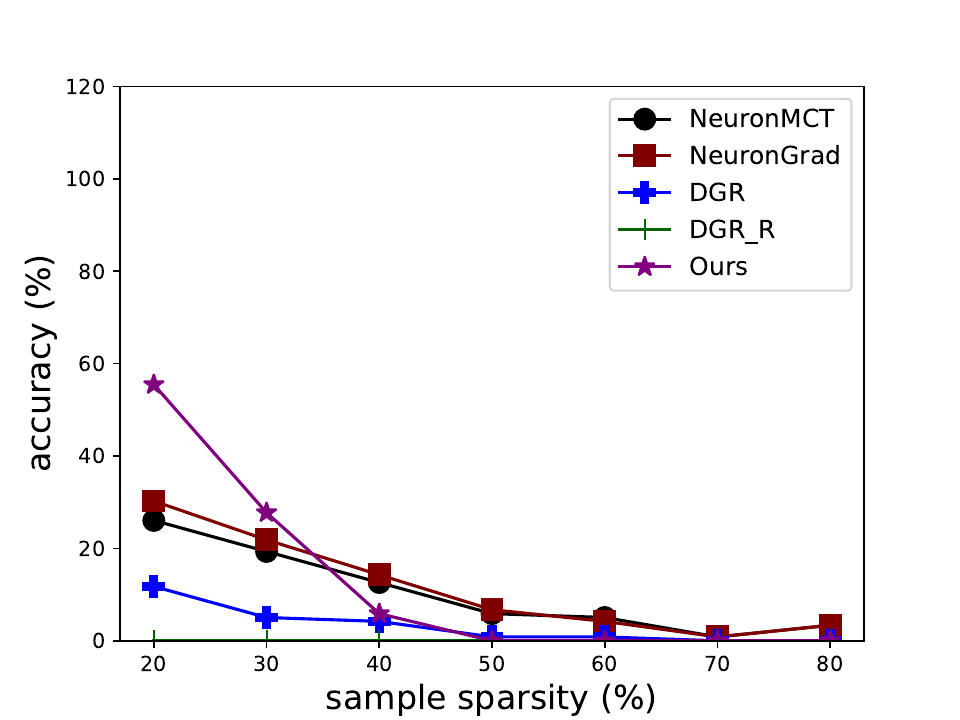}
  \caption{accuracy vs $\epsilon_{ss}$, $\epsilon_{cn} = 0.3$}
  \label{fig:aic_vs_ms}
\end{subfigure}
\caption{\textbf{Transferability Experiments on AlexNet \cite{alexnet} \& ImageNet \cite{imagenet}.} The mIC and ICr results are all 0. The mDC and accuracy are 0 when the $\epsilon_{cn} > 0.3$.}
\label{fig:transfer_mic_vs_cn_vgg_cifar10}
\end{figure*}

\begin{figure*}[h]
\begin{subfigure}{.24\textwidth}
  \centering
  \includegraphics[width=1.1\linewidth]{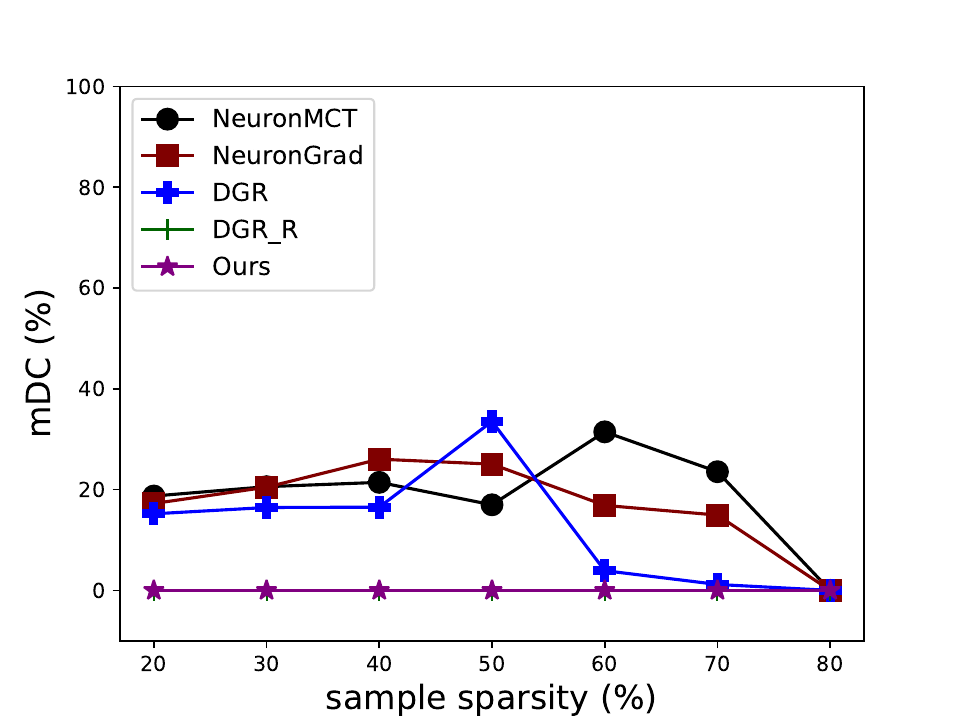}
  \caption{mDC vs $\epsilon_{ss}$, $\epsilon_{cn} = 0.1$}
  \label{fig:aic_vs_ms}
\end{subfigure}
\begin{subfigure}{.24\textwidth}
  \centering
  \includegraphics[width=1.1\linewidth]{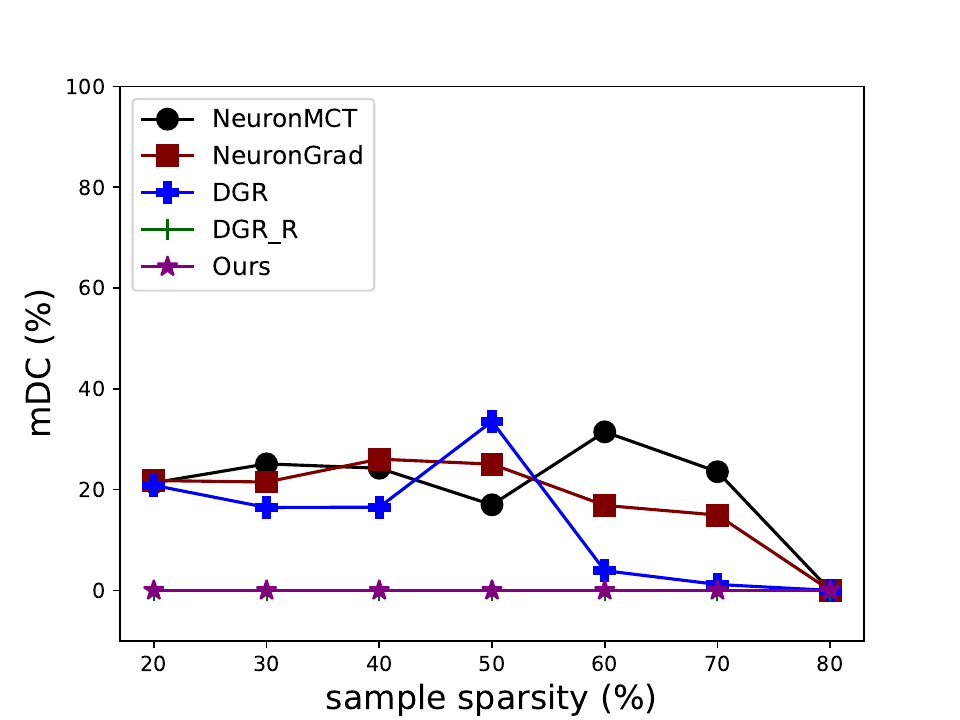}
  \caption{mDC vs $\epsilon_{ss}$, $\epsilon_{cn} = 0.2$}
  \label{fig:aic_vs_ms}
\end{subfigure}
\begin{subfigure}{.24\textwidth}
  \centering
  \includegraphics[width=1.1\linewidth]{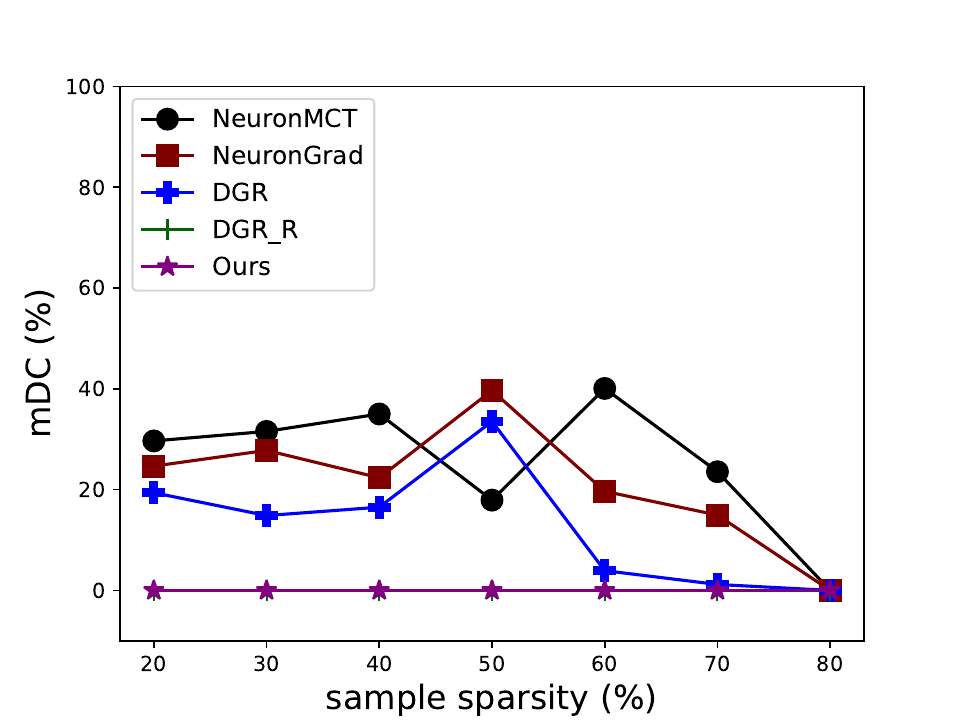}
  \caption{accuracy vs $\epsilon_{ss}$, $\epsilon_{cn} = 0.3$}
  \label{fig:aic_vs_ms}
\end{subfigure}
\begin{subfigure}{.24\textwidth}
  \centering
  \includegraphics[width=1.1\linewidth]{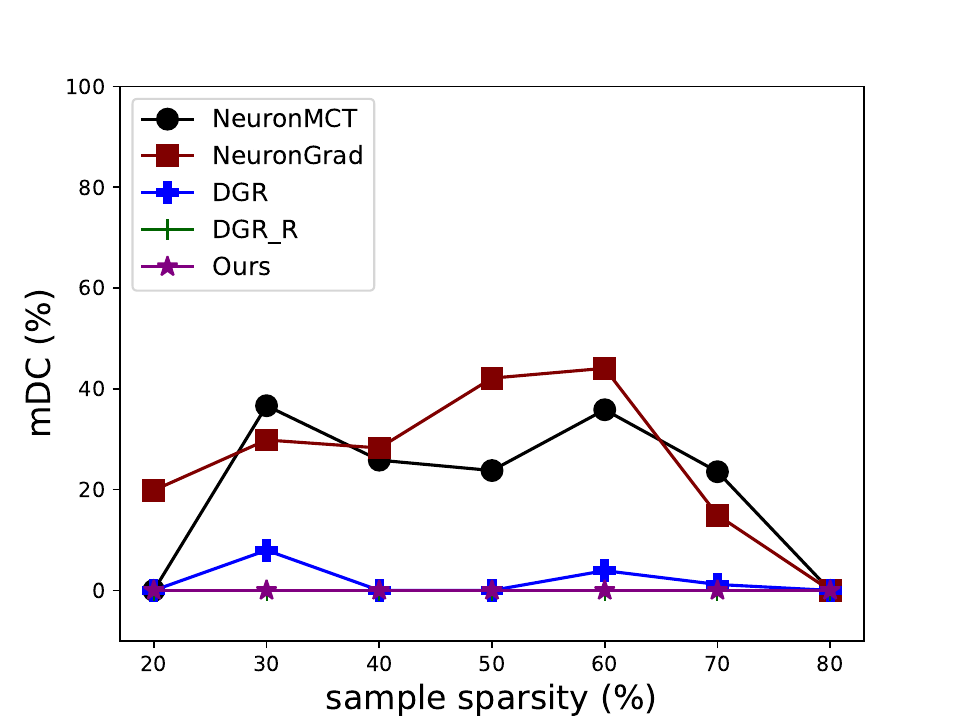}
  \caption{accuracy vs $\epsilon_{ss}$, $\epsilon_{cn} = 0.4$}
  \label{fig:aic_vs_ms}
\end{subfigure}
\caption{\textbf{Transferability Experiments on VGG-11 \cite{vgg} \& ImageNet \cite{imagenet}.} The mIC and ICr results are all 0. The mDC and accuracy are 0 when the $\epsilon_{cn} > 0.4$.}
\label{fig:transfer_mdc_vs_ss_vgg_image}
\end{figure*}

Fig. \ref{fig:transfer_mic_vs_ss_vgg_cifar10} illustrates the mIC vs different sample sparsities $\epsilon_{ss}$ for different class-relevant neural pathways firing sparsity $\epsilon_{cn}$ as defined in the paper. With increasing $\epsilon_{cn}$, the class-relevant neural pathways preserve less neurons, and the increasing $\epsilon_{ss}$ means that the class-relevant neural pathways are distilled from less samples of a given class. We see that our class-relevant neural pathways achieve better faithfulness than other baseline methods which contain less class-relevant neurons. Fig. \ref{fig:transfer_mic_vs_cn_vgg_cifar10} shows the mIC for increasing $\epsilon_{cn}$ for different sample sparsities $\epsilon_{ss}$. The class-relevant neural pathways of our method are consistently more faithful than baseline methods. Fig. \ref{fig:transfer_alexnet_image} and \ref{fig:transfer_mdc_vs_ss_vgg_image} show the mDC vs sample sparsity $\epsilon_{ss}$ for different $\epsilon_{cn}$. The ICr and mIC metrics are nearly 0 for all methods. The accuracy is also nearly 0 for all methods after $\epsilon_{cn} > 0.3$ on AlexNet and $\epsilon_{cn} > 0.4$ for VGG-11.

Table \ref{tab:trans_acc} gives the model performance, i.e. accuracy, of class-relevant neural pathways. We first observe that the accuracies of all methods are higher than the whole model which is 77.46\%. This could be due to that the neural pathways remove the noisy neurons that adversely impact the model's predictions. We also observe that the baselines NeuronMCT, NeuronGrad and DGR\_R have performance on par with our method even though the neurons in their class-relevant neural pathways are less class-relevant. We think that this is expected behavior because the whole model uses all neurons regardless of their class-agnostic/relevant characteristics, thus the class-agnostic neurons also contribute to the model's predictions. 

\begin{table}[!htp]\centering
\caption{Transferability accuracy with increasing sample sparsity $\epsilon_{ss}$ on AlexNet \cite{alexnet} and CIFAR-10 \cite{cifar10}.}\label{tab:trans_acc}
\scriptsize
\begin{tabular}{l | c | c | c | c| c| c }
\hline
Sample Sparsity $\epsilon_{ss}$ & 0.40 & 0.50 & 0.60 &0.70 &0.80 &0.9 \\
\hline
NeuronMCT \cite{PathwayGrad} &99.49 &99.00 &98.69 &98.74 &99.12 &99.29 \\
NeuronGrad \cite{PathwayGrad} &99.99 &99.96 &99.94 &99.93 &99.97 &99.98 \\
DGR \cite{dgr} &82.23 &82.02 &81.98 &81.94 &81.86 &82.08 \\
DGR\_R \cite{dgr} &100.00 &100.00 &100.00 &100.00 &100.00 &100.00 \\
\textbf{GEN-CNP (ours)} &100.00 &100.00 &100.00 &100.00 &100.00 &100.00 \\
\hline
\end{tabular}
\end{table}

\section{ROAP}
We report the additional results of the ROAP experiment on the VGG-11 model \cite{vgg} for dataset ImageNet \cite{imagenet} in Table \ref{tab:roap_vgg_imagenet}. The accuracy results of AlexNet \cite{alexnet} and VGG-11 \cite{vgg} models on CIFAR-10 are less than random guess.

\begin{table*}[!htp]\centering
\setlength{\extrarowheight}{0.5mm}
\setlength{\tabcolsep}{1.0mm}
\caption{ROAP on VGG-11 \cite{vgg} \& ImageNet \cite{imagenet}, accuracy metric (\%).}
\label{tab:roap_vgg_imagenet}
\small
\begin{tabular}{l|cccccccc}\toprule
path sparsity & 20 & 30 & 40 & 50 & 60 & 70 & 80 \\
\hline
Original Activation & \textbf{0.09} &0.11 &0.60 &80.33 &81.12 &81.42 &81.67 \\
NeuronMCT \cite{PathwayGrad} & 0.73 &0.64 &0.49 &0.54 &0.50 &0.62 &0.54 \\
NeuronIntGrad \cite{PathwayGrad} & 0.62 &0.55 &0.53 &0.70 &0.55 &0.65 &0.64 \\
DGR \cite{dgr} & 0.10 &0.29 &0.27 &0.26 &0.40 &0.37 &0.26 \\
DGR\_R \cite{dgr} & 89.03 &88.31 &88.94 &89.34 &88.96 &88.63 &88.85 \\
Ours & 0.17 &\textbf{0.07} &\textbf{0.10} &\textbf{0.12} &\textbf{0.11} &\textbf{0.13} &\textbf{0.16} \\
\bottomrule
\end{tabular}
\end{table*}

\section{Additional Qualitative Results}

For better visualizations, we train AlexNet \cite{alexnet} on a subset of 200 classes of ImageNet. The classes used are the same as in the Tiny ImageNet \footnote{https://github.com/DennisHanyuanXu/Tiny-ImageNet}. For each class, we randomly sample 500 images from the training dataset of the ImageNet \cite{imagenet}, and 50 images from the validation dataset of ImageNet \cite{imagenet}. 

\begin{figure*}[!t]
    \centerline{\includegraphics[width=1.0\textwidth]{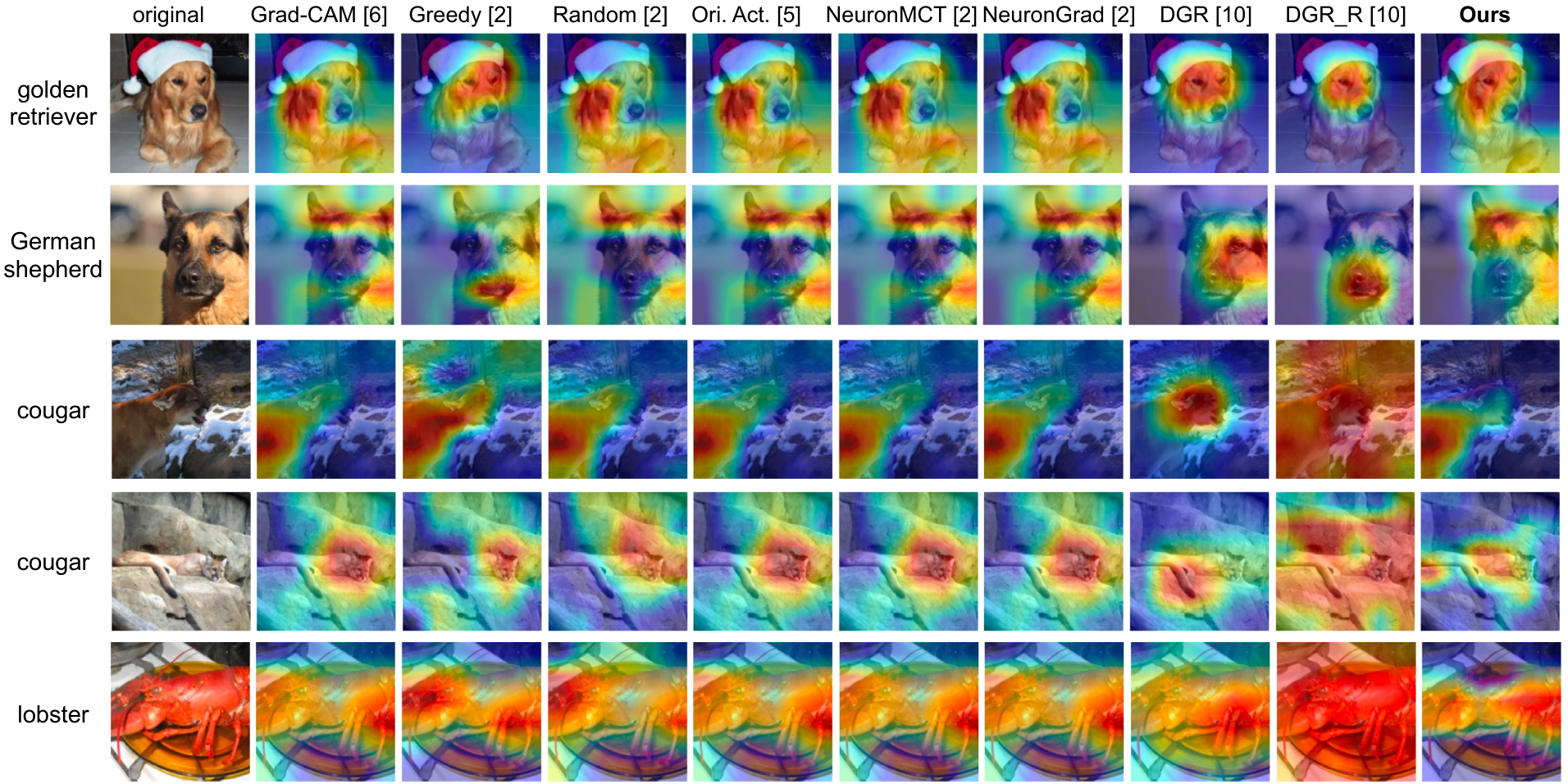}}
    \caption{\small{CAM Visualizations on ImageNet \cite{imagenet} using Grad-CAM. Our method compared to other methods seem to show less noise while focus more on the target animal.}}
    \label{fig:cam_image_vis_1}
\end{figure*}

\begin{figure*}[!t]
    \centerline{\includegraphics[width=1.0\textwidth]{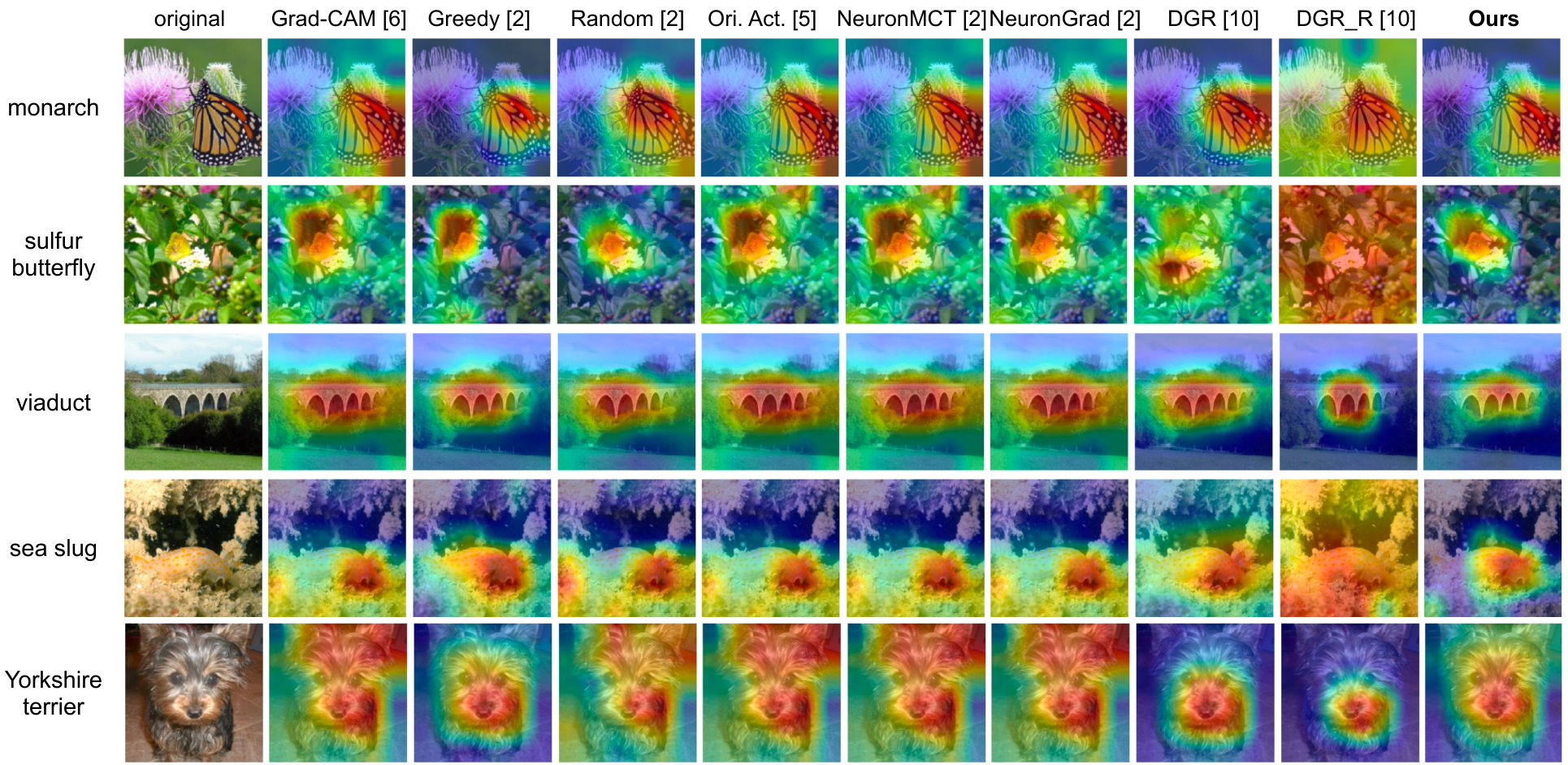}}
    \caption{\small{More CAM Visualizations on ImageNet \cite{imagenet} using Grad-CAM.}}
    \label{fig:cam_image_vis_2}
\end{figure*}

\begin{figure*}[!t]
    \centerline{\includegraphics[width=1.0\textwidth]{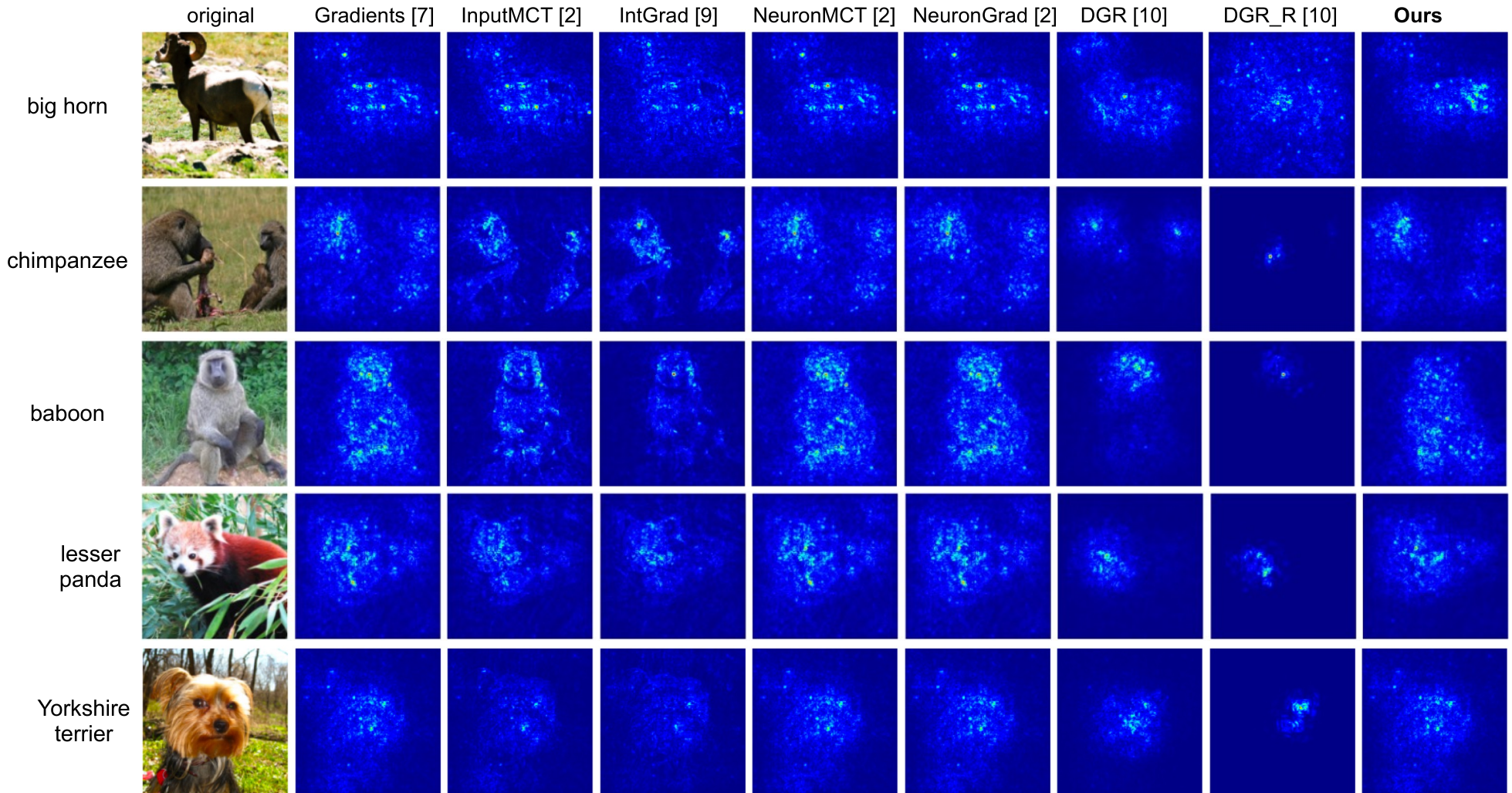}}
    \caption{\small{Neural Pathway Gradients Visualizations on ImageNet \cite{imagenet}. Our method compared to other methods seem to show less noise while have more salient features focus more on the target animal}}
    \label{fig:path_grad_vis_1}
\end{figure*}

\begin{figure*}[!t]
    \centerline{\includegraphics[width=1.0\textwidth]{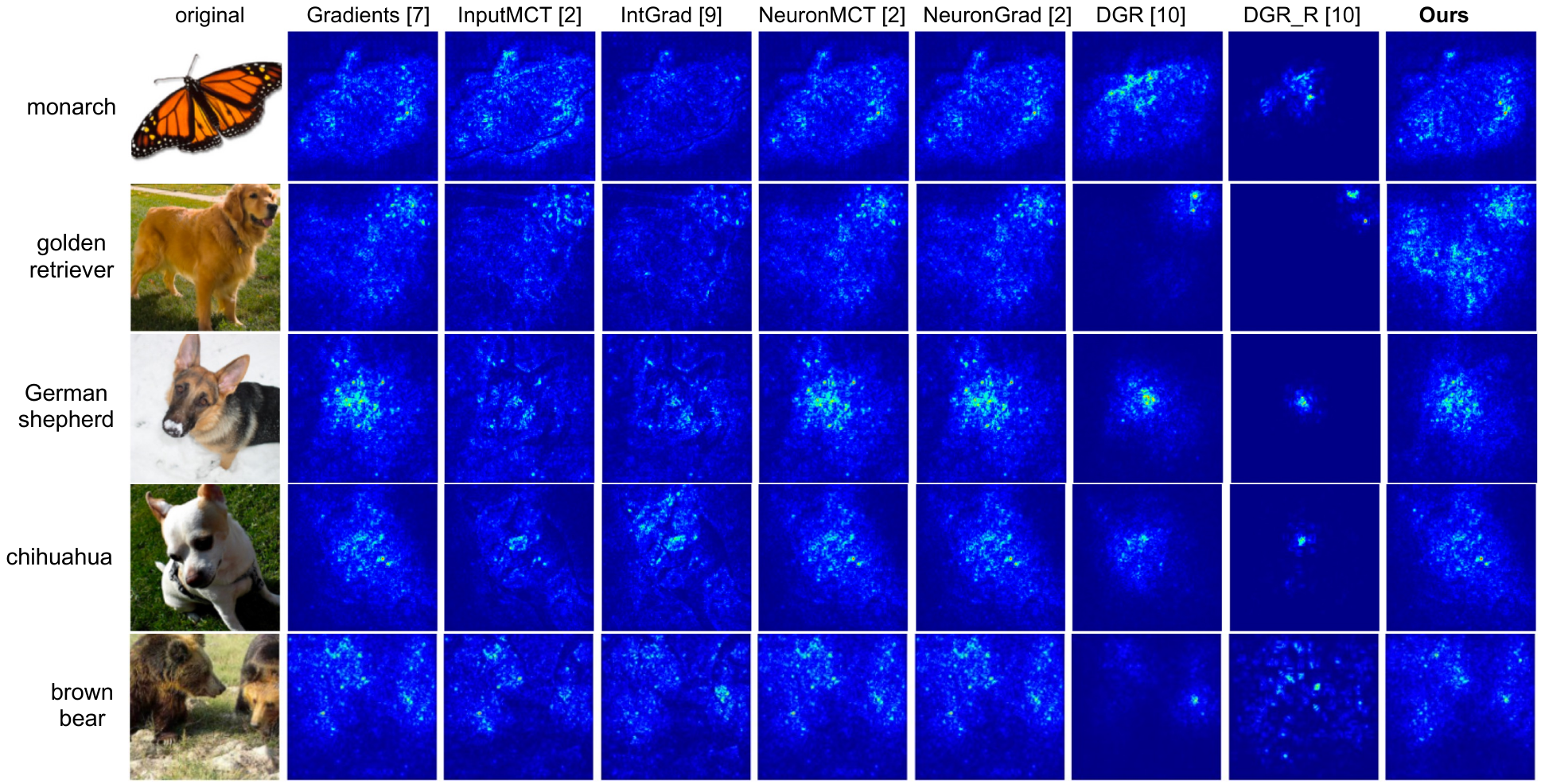}}
    \caption{\small{More Neural Pathway Gradients Visualizations on ImageNet \cite{imagenet}. }}
    \label{fig:path_grad_vis_2}
\end{figure*}

In Fig. \ref{fig:cam_image_vis_1} and \ref{fig:cam_image_vis_2}, we show the CAM visualizations of our method (last column) vs existing neural pathways methods, and Grad-CAM \cite{grad-cam}, and two pruning methods, \textit{Ori. Act.} which is based on magnitude pruning \cite{lecun_pruning} and \textit{Greedy} \cite{PathwayGrad}. In Fig. \ref{fig:path_grad_vis_1} and \ref{fig:path_grad_vis_2}, we show the gradients visualizations of our method (last column) vs existing neural pathways methods, and two backpropagation-based methods, IntGrad \cite{integrated} and Gradients \cite{gradient}.

\begin{figure*}[!t]
    \centerline{\includegraphics[width=1.0\textwidth]{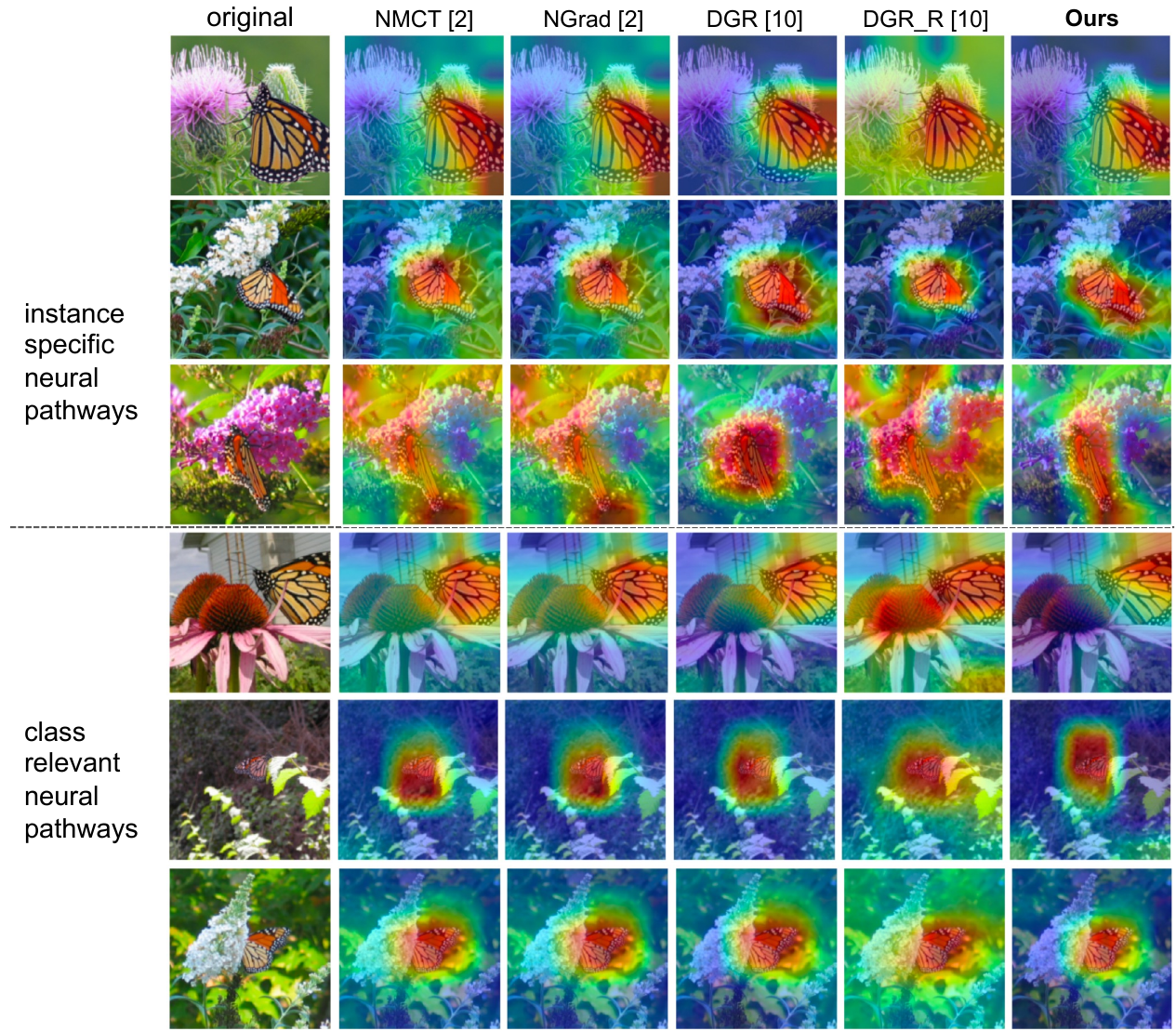}}
    \caption{\small{Neural Pathways Transferability Visualizations on ImageNet \cite{imagenet} using Grad-CAM. The class is monarch. The class-relevant neural pathways when transferred, focus more tightly on the body of the monarch butterfly than the background.}}
    \label{fig:transfer_vis_2}
\end{figure*}

\begin{figure*}[!t]
    \centerline{\includegraphics[width=1.0\textwidth]{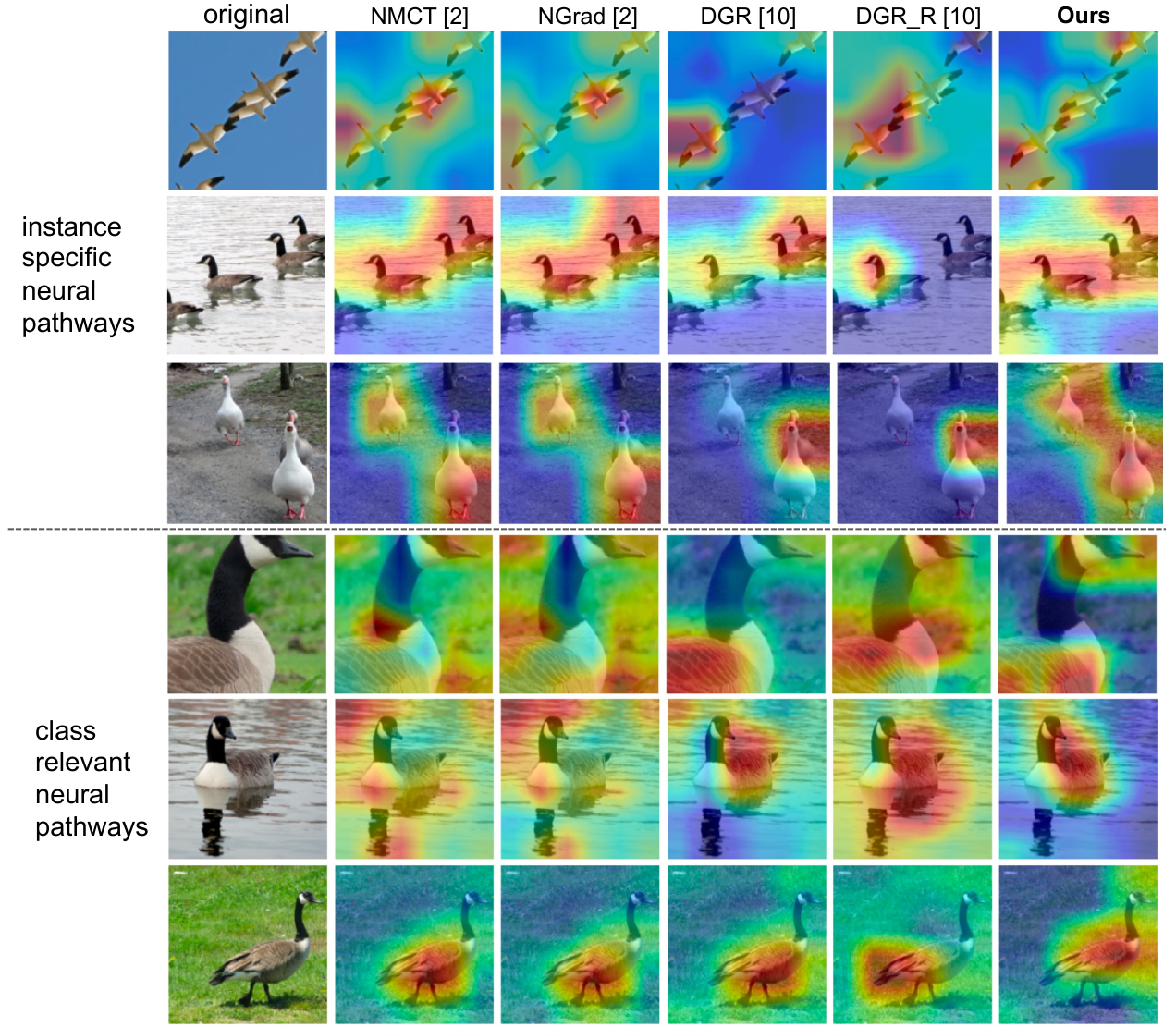}}
    \caption{\small{Neural Pathways Transferability Visualizations on ImageNet \cite{imagenet} using Grad-CAM. The class is goose. The class-relevant neural pathways when transferred, focus more tightly on the body of the goose than the background.}}
    \label{fig:transfer_vis_3}
\end{figure*}

In Fig. \ref{fig:transfer_vis_2} and \ref{fig:transfer_vis_3}, we show the visualizations of neural pathways transferability introduced in the paper. The top part contains three samples for which our method generates instance-specific neural pathways. Then, we build the class-relevant neural pathways from them by aggregating the class-relevant neurons.

\end{document}